\documentclass[times,twocolumn,final]{elsarticle} 
\usepackage{medima}
\usepackage{bbding}
\usepackage{framed}
\usepackage{latexsym}
\usepackage{url}
\usepackage{colortbl}
\usepackage{pifont}
\usepackage{xcolor}
\usepackage{cite}
\usepackage{makecell}
\usepackage{xspace}

\usepackage{amsmath,amssymb,amsfonts,bm}
\usepackage{algorithm}
\usepackage{algorithmicx}
\usepackage{algpseudocode}
\usepackage{setspace}
\usepackage{mathrsfs}
\usepackage{graphicx,epstopdf,subfigure}
\usepackage{multirow}
\usepackage{makecell}
\usepackage{tabularx,threeparttable}
\usepackage{textcomp}
\usepackage{booktabs}
\usepackage{diagbox}
\usepackage{setspace}
\usepackage{dsfont}
\usepackage{arydshln}
\usepackage{bbm}
\usepackage{soul,xcolor}
\usepackage{color}
\soulregister{\cite}{7}
\soulregister{\ref}{7} 
\soulregister{\citep}{7}
\soulregister{\citet}{7}
\soulregister{\pageref}{7}
\usepackage[colorlinks,linkcolor=blue]{hyperref}

\newcommand{\iou}{\texttt{IoU}\xspace}

\newcommand{\evda}{\texttt{EV17}\xspace}
\newcommand{\evdb}{\texttt{EV18}\xspace}
\soulregister{\iou}{7}
\soulregister{\evda}{7}
\soulregister{\evdb}{7}
\usepackage{tikz}
\usepackage{cleveref}
\definecolor{cr}{HTML}{E14591}
\definecolor{r1}{HTML}{FF6933}
\definecolor{r2}{HTML}{349B90}
\definecolor{r3}{HTML}{9579d1}
\definecolor{tc}{HTML}{E7EAEF}
\journal{Medical Image Analysis}
\begin{document}
\verso{\textit{XXX et~al.}}
\begin{frontmatter}
\title{LACOSTE: Exploiting stereo and temporal contexts for surgical instrument segmentation}%

\author[1,2]{Qiyuan \snm{Wang}}
\author[1,2]{Shang \snm{Zhao}}
\author[1,2]{Zikang \snm{Xu}}
\author[1,2,3,4]{S Kevin \snm{Zhou}\corref{cor1}}
\cortext[cor1]{Corresponding author: 
  Email: skevinzhou@ustc.edu.cn}

\address[1]{School of Biomedical Engineering, Division of Life Sciences and Medicine, University of Science and Technology of China (USTC), Hefei Anhui, 230026, China}
\address[2]{Center for Medical Imaging, Robotics, Analytic Computing \& Learning(MIRACLE), Suzhou Institute for Advanced Research, USTC, Suzhou, Jiangsu, 215123, China}
\address[3]{Key Laboratory of Precision and Intelligent Chemistry, USTC, Hefei Anhui, 230026, China}
\address[4]{Key Lab of Intelligent Information Processing of Chinese Academy of Sciences(CAS), Institute of Computing Technology, CAS, Beijing, 100190, China}

\begin{abstract}
Surgical instrument segmentation is instrumental to minimally invasive surgeries and related applications. Most previous methods formulate this task as single-frame-based instance segmentation while ignoring the natural temporal and stereo attributes of a surgical video. As a result, these methods are less robust against the appearance variation through temporal motion and view change. In this work, we propose a novel \textbf{LACOSTE} model that exploits \textbf{L}ocation-\textbf{A}gnostic \textbf{CO}ntexts in \textbf{S}tereo and \textbf{TE}mporal images for improved surgical instrument segmentation. Leveraging a query-based segmentation model as core, we design three performance-enhancing modules. Firstly, we design a disparity-guided feature propagation module to enhance depth-aware features explicitly. To generalize well for even only a monocular video, we apply a pseudo stereo scheme to generate complementary right images. Secondly, we propose a stereo-temporal set classifier, which aggregates stereo-temporal contexts in a universal way for making a consolidated prediction and mitigates transient failures. Finally, we propose a location-agnostic classifier to decouple the location bias from mask prediction and enhance the feature semantics. We extensively validate our approach on three public surgical video datasets, including two benchmarks from EndoVis Challenges and one real radical prostatectomy surgery dataset GraSP. Experimental results demonstrate the promising performances of our method, which consistently achieves comparable or favorable results with previous state-of-the-art approaches.

\end{abstract}
\end{frontmatter}
\section{Introduction}
Computer-assisted intervention (CAI) has emerged as a transformative force in surgical procedures as it enhances patient safety, improves operative quality, reduces adverse event, and shortens recovery period~\citep{maier2017surgical}. In this context, achieving semantic and instance segmentation of a surgical scene, as captured by surgical stereo cameras, plays a critical role in modern CAI systems. Semantic annotations enable cognitive assistance by providing pixel-wise contextual awareness of instruments, which is essential for supporting various downstream tasks, including surgical decision-making~\citep{loftus2020artificial,maier2022surgical}, surgical navigation~\citep{allan20202018}, and skill assessment~\citep{curtis2020association, liu2021towards}. Accurately identifying instruments and their spatial locations is a key focus in CAI, encompassing endeavors such as tool pose estimation~\citep{hein2021towards}, tool tracking and control~\citep{du2019patch}, and surgical task automation~\citep{nagy2019dvrk}. Moreover, the integration of surgical scene contexts can facilitate selective overlaying of different objects within augmented reality environments, opening new possibilities for the next generation of surgical education~\citep{allan20202018}. 

Consequently, there is a growing need for automated segmentation of surgical instruments, prompting active research in this domain. However, achieving precise instance segmentation from surgical videos confronts significant challenges. Complicated surgical scenes exhibit a low inter-class variance, such as variations between different instruments, and a high intra-class variance, such as instances of dynamic posed instruments. Class imbalance is also prevalent in surgical scenes, with the identification of small objects and rarely used instruments proving to be difficult. Challenges further arise from motion blur, lighting changes, and occlusions caused by smoke and blood~\citep{bouget2017vision}.
In recent years, remarkable progress has been made in the development of semantic and instance segmentation algorithms within the surgical community. While notable strides have been made in addressing this challenge, there are still some specific challenges that necessitate further improvement. Addressing these challenges represents crucial avenues for future research, aiming to advance instrument segmentation techniques and foster a more comprehensive understanding of surgical scenes \citep{wei2022FD}. This constitutes the focus of this paper, that is, improving the performance of \textbf{surgical instrument segmentation (SIS)}.

For SIS, the effectiveness of query-based segmentation~(QBS) methods has been validated by previous works. The findings of \citep{baby2023forks} indicate that instrument misclassification primarily leads to low performance of current SIS methods. While most methods yield satisfactory results in terms of both the bounding box and segmentation mask, they frequently misclassify the output box or mask. Therefore, we follow the QBS paradigm and derive our method, grounded on three limitations of current methods, to mitigate instrument misclassification for surgical scenario. Firstly, surgical frames are recorded by stereo cameras and have inherent stereo attributes that are largely ignored by previous works. The complementary depth-aware information from stereo frame can enhance current view features, which help instrument localization and recognition for complex scenes. From this, we design a \textit{disparity-guided feature propagation}~(DFP) module, which integrates features from stereo views with an offline disparity estimation network explicitly. DFP can be inserted into QBS baseline without additional trainable parameters. To generalize well to a monocular setting, we also propose a pseudo stereo mechanism, which generates complementary right frames within multiple disparity ranges. Some examples are illstrated in Figure~\ref{fig:feature_propagation}.

Secondly, most QBS methods for SIS group image pixels into different segments including binary masks and corresponding instrument categories from only frame-wise information. However, the appearance of instruments across temporal and stereo frames exhibits great variations due to motion blur, occlusion, and so on. Hence, we address SIS from a tracklet perspective and propose a \textit{stereo-temporal set classifier} (STSCls), which decides the final instrument category of each segment after aggregating instrument information through a stereo clip. A tracklet means a short track of instance across frames. In this paper, we extend the tracklet definition to \textit{a set of object query embeddings corresponding to the same identity across temporal and stereo dimensions}, wherein each item  of tracklet represents an object query embedding. Since STSCls takes stereo-temporal contexts within a tracklet into account for each instrument segment, it is robust against temporal shifts and view changes and avoids a collapse of final predictions from transient failures. STSCls can be cascaded on top of QBS baseline and trained jointly with tracklets generated from QBS outputs. Given the absence of video instance annotations for most surgical datasets, we design a query alignment mechanism and identity alignment loss to align identity IDs based on query indexes instead of an additional track head for tracklets generation. Due to the generality of these mechanisms, STSCls can be extended to train with various tracklet configurations including stereo clip and monocular clip.

Finally,  binary mask and instrument category of each segment for QBS methods are originated from the same object query~(embedding).
The embeddings learn not only semantic contents but also location biases, which can make them scattered into different clusters in semantic feature space and hence have a negative influence for classification. In fact, the features used for segmentation are not necessarily appropriate for classification. To this end, we propose an additional \textit{location-agnostic classifier}~(LACls) to decouple the location bias from semantic information. With a plug-and-play design, this classifier can be easily plugged on top of QBS architecture. In this work, LACls receives the mask prediction from QBS baseline and cropped images to extract features. The features of location-agnostic structure are clustered compactly within the same category.

We extensively evaluate our method on three publicly available surgical datasets, including two benchmarks from EndoVis Challenges together with one real radical prostatectomy surgery dataset GraSP and demonstrate that our new approach achieves comparable or favorable results with existing state-of-the-art (SOTA). Our main contributions are:
\begin{enumerate}
\item
We propose a novel query-based segmentation framework, \textbf{LACOSTE}, which jointly exploits Location-Agnostic COntexts in Stereo and TEmporal images for improved surgical instrument segmentation. 
\item
We design a disparity-guided feature propagation~(\textbf{DFP}) module to enhance current feature with stereo cue, a stereo-temporal set classifier~(\textbf{STSCls}) to improve semantic inference with temporal-stereo context, and a cascaded location-agnostic classifier~(\textbf{LACls}) to mitigate the negative influence of location bias for classification.
\item
We develop a query alignment mechanism and identity alignment loss to promote instance consistency across frames in one stereo clip instead of a track head.
\item
Our method achieves performance gains in three open benchmark datasets when compared with prior state-of-the-art methods.
\end{enumerate}
\section{Related Work}
\underline{CNN-based SIS.} Initial efforts in surgical robotics community are based on Convolutional Neural Networks (CNNs). For instance, TernausNet~\citep{shvets2018automatic} proposes a pretrained U-Net~\citep{ronneberger2015u} model for the segmentation of a restricted variety of surgical instruments. U-NetPlus modifies a U-Net architecture and data augmentation strategies to improve performance~\citep{hasan2019u}. PAANet pays more attention to multi-scale attentive features~\citep{ni2020pyramid}. To improve the segmentation results, some researchers incorporate additional priors such as optical flow and motion cue~\citep{zhao2020learning, jin2019incorporating}, stereoscopic information~\citep{mohammed2019streoscennet}, or saliency maps~\citep{islam2019learning}. These previous models formulate SIS as a single frame per-pixel classification task, which often produces disconnected areas and ignores a multi-instance nature. Some works have argued for formulating the task as multi-class instance segmentation in a single frame. ISINet~\citep{gonzalez2020isinet} applies a Mask-RCNN~\citep{he2017mask} accompanied with optical flow to address instance-based instrument segmentation. S3Net~\citep{baby2023forks} analyzes the low-IoU performance of previous methods and pays more attention to classification. Specially, it applies a mask-based attention and contrastive loss to address the variation in aspect ratio and inter-class appearance similarity.

\underline{Transformer-based SIS.} In recent years, Vision Transformers (ViTs)~\citep{liu2021swinv2,dosovitskiy2020image} have emerged as the leading models in diverse computer vision tasks, showcasing their state-of-the-art performance. Among these works, query-based segmentation architectures like MaskFormer~\citep{cheng2021per}, which utilize a fixed-size set of learnable object queries to predict regions, have tremendous advantages for universal segmentation tasks. The effectiveness of treating instance segmentation as a joint problem of mask segmentation and mask classification has been validated in many segmentation tasks. Inspired from these works, some researchers introduce Transformers into SIS task. In this regard, ~\citep{sun2022parallel} is the first to integrate Transformers into instrument segmentation by combining CNNs with Swin Transformers as backbone. \citep{dhanakshirur2023learnable} reveal that incorrect query initialization limits the effectiveness of current query-based methods and propose a class-agnostic Query Proposal Network (QPD) to improve query initialization.

\underline{Temporality-enhanced SIS.} Temporal attributes have been taken into account in SIS tasks. Previous CNN-based approaches often incorporate optical flow information without fully exploiting global video contexts. Compared with CNN-based methods, video Transformers have advantages of aggregating long-term video reasoning and showcase their superior performance in video tasks. TraSeTr~\citep{zhao2022trasetr} is a track-to-segment Transformer that dynamically integrates tracking cues to assist instance-level surgical instrument segmentation. STSwinCL~\citep{jin2022exploring} employs a video Swin Transformer with a contrastive learning approach for panoptic segmentation of surgical scenes in videos. TAPIR~\citep{valderrama2022towards} introduces a video Transformer-based model for multi-level surgical workflows analysis accompanied with Deformable DETR~\citep{zhu2020deformable} for instrument detection, offering an improved solution for SIS. TAPIS~\citep{ayobi2023towards} is a multi-task transformer-based architecture that combines a global video feature extractor with a localized region proposal network for actions, phases, steps recognition and instrument segmentation.CaRTS~\citep{ding2023rethinking} introduces temporal constraints on kinematics data for counterfactual surgical segmentation. MATIS~\citep{ayobi2023matis} utilizes pixel-wise attention for targeting instrument areas and video transformers for temporal information.

\underline{Set Classifier.} Recently, it has been shown that making a consolidated decision from multiple frames is beneficial to improve the quality of detection and segmentation in natural scene~\citep{bertasius2020classifying,hwang2021video,ke2021prototypical}. SCTrack \citep{hwang2022cannot} first pays more attention to using set classifier to improve the classification performance. SCTrack uses a track head to compose boxes tracklets dynamically and use set classifier with region proposals features to predict the final results of instances. The effectiveness of this structure is based on extensive region proposals that explain the same instance from different views generated by Region Proposal Network module and sequence generation strategy.

\begin{figure*}[!htbp]
    \centering
	\includegraphics[width=\textwidth]{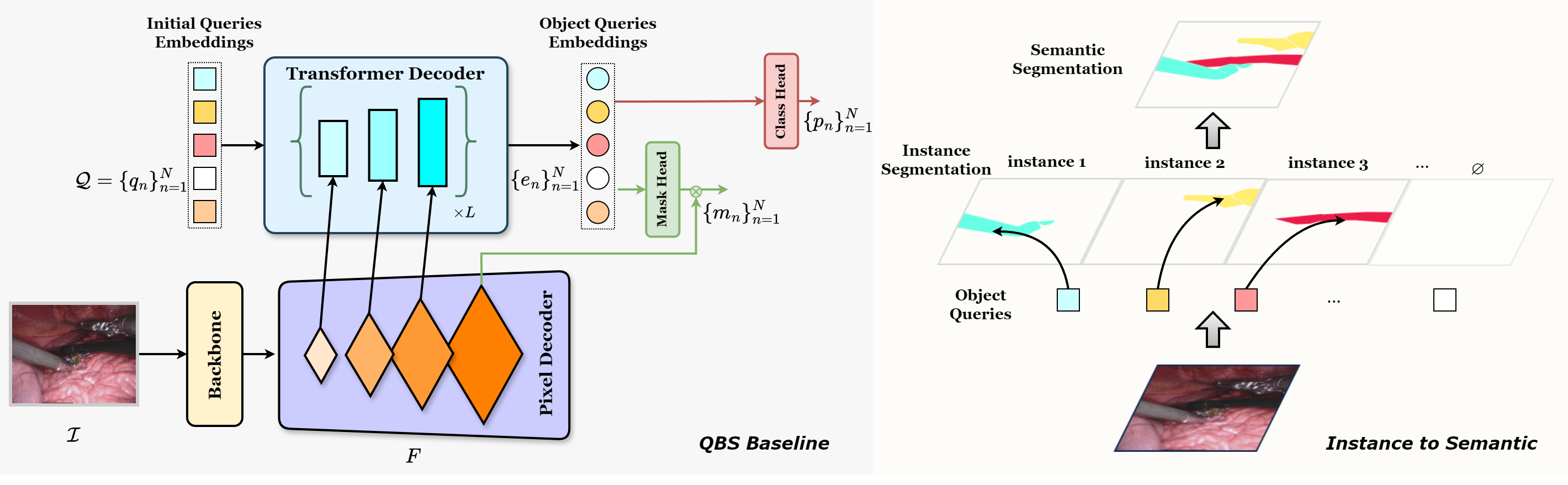}
	\caption{Schematic illustration of Query-Based Segmentation architecture. The key components include backbone encoder, pixel decoder and transformer decoder together with learnable queries(embeddings).}
	\label{fig:qbsa}
 \vspace{-3mm}
\end{figure*}

\begin{table*}[!ht]
\begin{center}
    \footnotesize
    \caption{The mathematical symbols in this work.}
    \renewcommand\arraystretch{1.2}
    \begin{tabular}{cll|cll|cll}
        \hline
        \multicolumn{3}{c|}{\textbf{QBS}} & \multicolumn{3}{c|}{\textbf{Frame Step}} & \multicolumn{3}{c}{\textbf{Tracklet Step}} \\ \hline
        Notions & Details & Description & Notions & Details & Description & Notions & Details & Description \\ \hline
        $\mathcal{I}$ & & input image   & $\mathcal{I^L}$ & & left view & $s$ & $\{s_n\}_{n=1}^N$ & tracklets \\
        $F$ & & pixel-wise features   & $\mathcal{I^R}$ & & right view  & $p^s$ & $\{p^s_n\}_{n=1}^N$ & class probability \\
        $n$ & & query index     & $t$ & & frame index & $e^s$ & $\{e^s_n\}_{n=1}^N$ & object query  \\
        \cline{7-9}
        $N$ & & queies number   & $t^*$ & & current time & \multicolumn{3}{c}{\textbf{L-Agnostic Step}} \\
        \cline{7-9}
        $\tilde{N}$ & & real instance number & $T$ & & clip length & $p^a$ & $\{p^a_n\}_{n=1}^N$ & class probability \\
        $p$ & $\{p_n\}_{n=1}^N$ & class probability & $\mathcal{C}$ & $\{(\mathcal{I^{L}}(t), \mathcal{I^{R}}(t))\}_{t=1}^{T}$ & stereo clip & $e^a$ & $\{e^a_n\}_{n=1}^N$ & object query \\
        $m$ & $\{m_n\}_{n=1}^N$ & binary mask & $p^b$ & $\{p^b_n\}_{n=1}^N$ & class probability & & & \\
        \cline{7-9}
        $e$ & $\{e_n\}_{n=1}^N$ & object query & $m^b$ & $\{m^b_n\}_{n=1}^N$ & binary mask &  \multicolumn{3}{c}{\textbf{Final Output}}  \\
        \cline{7-9}
        $\mathcal{Q}$ & $\mathcal{Q}=\{q_n\}_{n=1}^N$ & initial query & $e^b$ & $\{e^b_n\}_{n=1}^N$ & object query & $p^f$ & $\{p^f_n\}_{n=1}^N$ & class probability \\
        $z$ & $z = \{(p_n, e_n, m_n)\}_{n=1}^N$ & prediction outputs   & & & & $m^f$ & $\{m^f_n\}_{n=1}^N$ & binary mask  \\
        $\tilde{z}$ & $\tilde{z} = \{(\tilde{c_{n}}, \tilde{m_{n}})\}_{n=1}^{\tilde{N}}$ & ground truth & & & & & &  \\
        \hline
    \end{tabular}
    \label{tab:symbol}
\end{center}
\vspace{-5mm}
\end{table*}

\section{Query-Based Segmentation}
\subsection{QBS Preliminaries}

The QBS method partitions image pixels into $N$ segments by predicting $N$ binary masks and $N$ corresponding category labels, where $N$ is significantly larger than the real segment number $\tilde N$. QBS represents each segment with a feature vector (``object query embedding”) which can be processed into category label and binary mask. The key challenge is to find good representations for each segment. For simplicity, we term object query embedding as object query in the following. As concluded in~\citep{cheng2021per}, a meta QBS architecture would be composed of three components: backbone, pixel decoder, and Transformer decoder together with trainable queries.

Given an image $\mathcal{I}$, the backbone $\theta$ first extracts low-resolution features. Then, the pixel decoder $\delta$ that gradually upsamples low-resolution features from the backbone to generate high-resolution per-pixel features $F$. Finally, the Transformer decoder $\zeta$ together with $N$ initial learnable query embeddings $\mathcal{Q} =\{q_n\}_{n=1}^{N}$ operates on image features to process $N$ object query embeddings $\{e_{n}\}_{n=1}^{N}$. The final predictions of $N$ segments are a set of $N$ probability-embedding-mask pairs $z = \{(p_n, e_n, m_n)\}_{n=1}^N$, where the probability distribution $p_n$ contains $C$ classes and an auxiliary “no object” label~($\varnothing$) to denote segments that do not correspond to any classes; $e_n$ is object query embedding to represent segment, where $n$ means query index; and $m_n$ is binary mask. The object query assigned with “no object” label is termed as a \textit{non-object query}. In this research, we \textbf{formulate each segment as instance and merge the selected instances into one semantic segmentation map}. The overall pipeline can be illustrated in Figure~\ref{fig:qbsa}.

The training losses are composed of classification and binary segmentation parts as represented in Eq.~(\ref{eq:qbsa}), where $\lambda_{\mathrm{bce}}$, $\lambda_{\mathrm{dice}}$ and $\lambda_{\mathrm{cls}}$ are weighted hyper-parameters. Hungarian matching makes instances matching between the predictions and ground truths $\tilde{z} = \{(\tilde{c_{n}}, \tilde{m_{n}})\}_{n=1}^{\tilde{N}}$ ($\tilde{c_{n}}$/$\tilde{m_{n}}$ means class/binary mask ground truth). The cross entropy~(\textbf{CE}) loss $\mathcal{L}_{cls}$ is applied for classification. For binary segmentation, the joint losses include binary cross entropy~(\textbf{BCE}) loss $\mathcal{L}_{bce}$ and the Dice loss $\mathcal{L}_{dice}$.

\begin{gather}
   \mathcal{L}_{cls} = \sum_{n=1}^{N}\mathbf{CE}(p_n,\tilde{c_n}), \notag \\
   \mathcal{L}_{bce} = \sum_{n=1}^{\tilde{N}} \mathbf{BCE}(m_n,\tilde{m_n}), \mathcal{L}_{dice} = \sum_{n=1}^{\tilde{N}} \mathbf{Dice}(m_n,\tilde{m_n}) , \notag \\
   \mathcal{L}_{baseline} = \lambda_{\mathrm{bce}}\mathcal{L}_{bce} + \lambda_{\mathrm{dice}}\mathcal{L}_{dice} + \lambda_{\mathrm{cls}}\mathcal{L}_{cls}.   \label{eq:qbsa}
\end{gather}

\subsection{Limitations of Current QBS}
However, the original QBS framework \emph{does} have limitations when applied to the SIS tasks. Firstly, current methods for SIS disregard additional information from stereo view  which can enhance the precision of instrument localization and recognition. Secondly, 
prior studies typically focus on frame-wise predictions, neglecting the temporal characteristics of surgical videos, thereby leading to transient failures. Finally, QBS methods derive binary mask and instrument category of each segment from the same object query, wherein inherent location biases detrimentally impact instrument recognition. These limitations motivate us to propose the LACOSTE model, which follows the QBS paradigm~(Mask2Former) and improves the effectiveness of current SIS works accompanied with three proposed modules including DFP, STSCls and LACls.

\label{sec:method}
\begin{figure*}[!htbp]
	\centering
	\includegraphics[width=\textwidth]{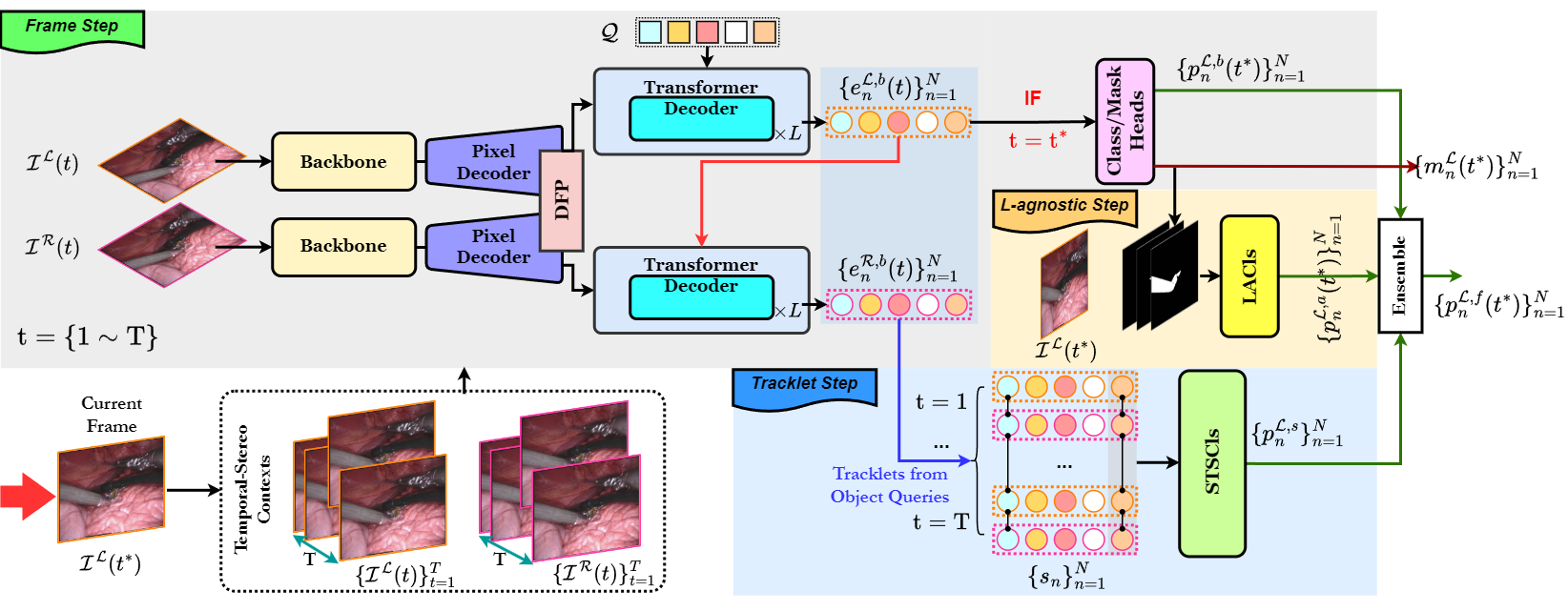}
    \caption{Schematic illustration of an overview pipeline of our LACOSTE method. Specifically, the prediction flow is divided into Frame step, Tracklet step, and L-Agnostic step. Together with proposed DFP, STSCls and LACls, all steps improve SIS from different perspectives.}
	\label{fig:overview}
\end{figure*}

\section{The LACOSTE Method}

\subsection{The Three Inference Steps}

LACOSTE adheres to the QBS paradigm in conjunction with three proposed modules: DFP, STSCls, and LACls. The overall inference pipeline of LACOSTE can be divided into three steps which aim to enhancing mask \textbf{classification} ability from various perspectives as illustrated in Figure~\ref{fig:overview}. For \textbf{\textcolor{red}{Frame Step}}, the QBS baseline with DFP~(termed as \textbf{BDFP}) explores depth-aware information from stereo views and makes frame-wise prediction for each timestamp. For \textbf{\textcolor{red}{Tracklet Step}}, STSCls aggregates temporal or stereo contexts contained in tracklets which are composed of object queries from frame step and makes a consolidated tracklet prediction. For \textbf{\textcolor{red}{L-Agnostic Step}}, LACls extracts instrument content information decomposed with location biases and predicts a location-agnostic prediction. The \textbf{ensemble} results from three steps replace original QBS classification results as the final classification predictions and final binary mask predictions keep same as those from frame step.

We define some mathematical notions in Table~\ref{tab:symbol} and formulate the inference process of LACOSTE for current frame $\mathcal{I^L}(t^*)$ as Algorithm~\ref{alg:lacoste_algo}. The upper suffix for three steps and final outputs are represented by $b$, $s$, $a$ and $f$, respectively. Given a current frame $\mathcal{I^L}(t^*)$ within a stereo clip $\{(\mathcal{I^{L}}(t), \mathcal{I^{R}}(t))\}_{t=1}^{T}$ (where $t^*$ means current time, $T$ means clip length, $t^* \in [1,T]$), LACOSTE goes through three steps and outputs the final segmentation results as follow.

\begin{figure*}[!htbp]	\centering
	\includegraphics[width=\textwidth]{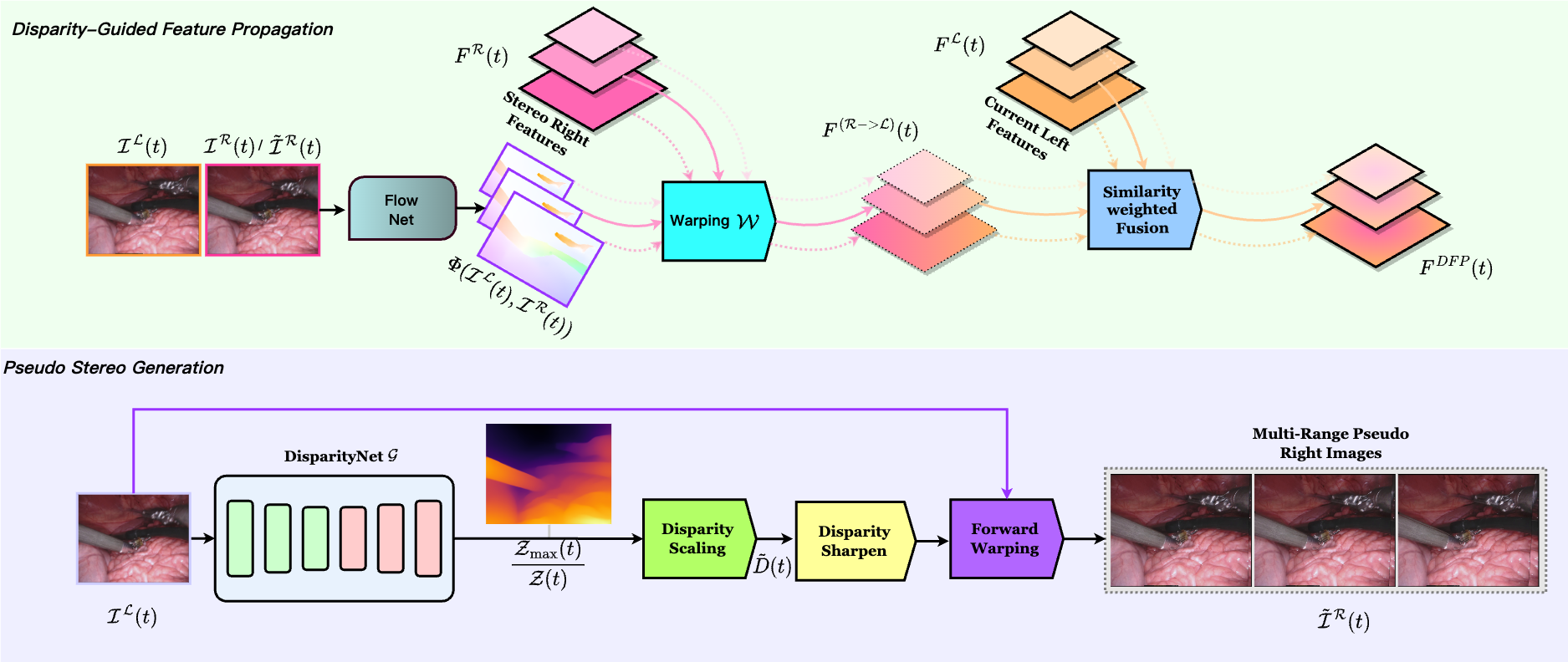}
	\caption{Schematic illustration of our DFP module and pseudo stereo generation mechanism. DFP fuses complement information from stereo pairs at feature level. Pseudo stereo generation mechanism generates virtual multi-range right images for only monocular dataset.} 
	\label{fig:feature_propagation}
\end{figure*}

\begin{algorithm}[!htb]
\caption{LACOSTE Inference Algorithm}
\label{alg:lacoste_algo}
\setstretch{1.02}
\textbf{Input}: Current Frame/Time $\mathcal{I^{L}}(t^{*})$/$t^*$, Corresponding Stereo Clip $\{(\mathcal{I^{L}}(t), \mathcal{I^{R}}(t))\}_{t=1}^{T}$, Clip Length $T$, Ensemble Weighted Parameters $\alpha_b$, $\alpha_s$, and $\alpha_a$. \\
\textbf{Params}: BDFP $\mathbf{\Phi_{B}}$,~STSCls $\mathbf{\Phi_{S}}$,~LACls $\mathbf{\Phi_{A}}$, Initial Queries $\mathcal{Q}$.\\
\textbf{Output}: ~Segmentation results $\mathcal{S}(t*) = \{(p^{f}_{n}(t*),m^{f}_{n}(t*))\}_{n=1}^{N}$.\\
\vspace*{-4mm}
\begin{algorithmic}[1]
\Statex \textbf{\textcolor{red}{Frame Step}}
\For{$t \leftarrow 1$ to $T$}
\If{$t > 1$}
\State $\mathcal{Q} = \{e_n^{b,\mathcal{L}}(t-1)\}_{n=1}^{N}$ \Comment{Query Alignment}
\Else
\State $\mathcal{Q} = \{q_n\}_{n=1}^{N}$
\EndIf
\If{$t = t^*$}
\State $\mathbf{\Phi_{B}}(\mathcal{I^{L}}(t), \mathcal{I^{R}}(t)) \xrightarrow{\mathcal{Q}} \bigcup\limits_{\mathcal{\tiny L,R}} \{(e_n^{b}(t),p_n^{b}
(t),m_n(t))\}_{n=1}^{N}$
\Else
\State $\mathbf{\Phi_{B}}(\mathcal{I^{L}}(t), \mathcal{I^{R}}(t)) \xrightarrow{\mathcal{Q}} \bigcup\limits_{\mathcal{\tiny L,R}} \{(e_n^{b}(t)
)\}_{n=1}^{N}$
\EndIf
\EndFor
\Statex \textbf{\textcolor{red}{Tracklet Step}}
\Statex Generate tracklets $\{s_n\}_{n=1}^{N}$ from object queries, where $s_n=\{(e_n^{b,\mathcal{L}}(t),e_n^{b,\mathcal{R}}(t))\}_{t=1}^T$
\For{$n \leftarrow 1$ to $N$}
\State $\mathbf{\Phi_{S}}(s_n) \rightarrow (e_n^s, p_n^s)$
\EndFor
\Statex \textbf{\textcolor{red}{L-Agnostic Step}}
\State $\mathbf{\Phi_{A}}(\mathcal{I^{L}}(t^{*}),\{m_n^{\mathcal{L}}(t^{*})\}_{n=1}^{N}) \rightarrow \{(e_n^a(t^{*}), p_n^a(t^{*}))\}_{n=1}^{N}$
\Statex \textbf{Final Results}
\Statex binary mask~$m_n^f(t*)=m_n(t*)$
\Statex class prediction $p_n^f(t*)= \alpha_{b} p_n^b(t*) + \alpha_{s} p_n^s + \alpha_{a} p_n^a(t*)$
\Statex \textbf{return} $\mathcal{S}(t*) = \{(p_{n}^f(t*),m_{n}^f(t*))\}_{n=1}^{N}$.
\end{algorithmic}
\end{algorithm}

\begin{enumerate}
    \item[$\diamond$] \textbf{\textcolor{red}{Frame Step.}}
    For each timestamp, BDFP receives a pair of stereo frames $(\mathcal{I^{L}}(t), \mathcal{I^{R}}(t))$ and derives the frame-wise predictions. Query alignment operation is applied to align identities across different timestamps. As shown in Eq.~(\ref{eq:frame_wise}), BDFP outputs probability-embedding-mask pairs at current time $t^*$ while it outputs only object query embeddings at other timestamps.
    \begin{equation}
    \label{eq:frame_wise}
    \left\{
    \begin{array}{ccl}
    \bigcup\limits_{_\mathcal{L,R}} \{(e_n^{b}(t),p_n^{b}(t),m_n(t))\}_{n=1}^{N} & , & t=t^*, \\
    \bigcup\limits_{_\mathcal{L,R}} \{(e_n^{b}(t))\}_{n=1}^{N} & , & t\neq t^*.
    \end{array}
    \right.
    \end{equation}
    \item[$\diamond$] \textbf{\textcolor{red}{Tracklet Step.}} We collect object queries through a stereo clip from frame step and generate tracklets $\{s_n\}_{n=1}^{N}$ (where $s_n=\{(e_n^{b,\mathcal{L}}(t),e_n^{b,\mathcal{R}}(t))\}_{t=1}^T$) based on query index $n$. STSCls receives tracklets and makes tracklet-level class prediction $\{(e_n^s, p_n^s)\}_{n=1}^N$.
    \item[$\diamond$] \textbf{\textcolor{red}{L-Agnostic Step.}} LACls receives processed images based on binary masks $\{m_n^{\mathcal{L}}(t^{*})\}_{n=1}^{N}$ from frame step and outputs location-agnostic predictions $\{(e_n^a(t^{*}), p_n^a(t^{*}))\}_{n=1}^{N}$. 
\end{enumerate}

The final outputs of LACOSTE for current frame are composed of binary masks $\{m_{n}^{\mathcal{L}}(t^*)\}_{n=1}^{N}$ and corresponding class predictions $\{p_{n}^{f,\mathcal{L}}(t^*)\}_{n=1}^{N}$. The class predictions are ensemble with frame class prediction $p^{b}$, tracklet class prediction $p^{s}$ and location-agnostic class prediction $p^{a}$. Considering the outputs of the Frame step are utilized in the other two steps, LACOSTE first performs the Frame step and subsequently executes the Tracklet and L-Agnostic steps in parallel. We will elaborate all three modules and training losses in the following sub-sections.

\subsection{Disparity-Guided Feature Propagation (DFP)}
\textbf{DFP Structure \textsc{\&} Formulation.} As presented in the top row of Figure~\ref{fig:feature_propagation}, DFP makes a bridge to exploring stereo information at feature level inspired by temporal tasks~\citep{zhao2022trasetr,zhu2017flow}. It converts the stereo right features to current left features guided with an offline flow estimation model and fuses weighted pair features adaptively. Without inducing additional trainable parameters, this operation can enhance current left features with stereo right features. Specifically, LACOSTE first extracts high-resolution per-pixel feature maps $(F^{\mathcal{L}}(t),F^{\mathcal{R}}(t))$ for a given stereo pair $(\mathcal{I^{L}}(t),\mathcal{I^{R}}(t))$, respectively. An offline optical-flow network $\Phi$ is introduced to estimate disparity between stereo frames. The DFP module applies a backward warping function $\mathcal{W}$ to make the stereo features $F^{\mathcal{R}}(t)$ aligned with the current features $F^{\mathcal{L}}(t)$. After that, the warped stereo features are fused adaptively with current features. The below equations represent this operation.
\begin{equation}
    F^{(\mathcal{R}\rightarrow\mathcal{L})}(t) = \mathcal{W}(F^{\mathcal{R}}(t), \Phi(\mathcal{I^{L}}, \mathcal{I^{R}})),
    \label{eq:propogation}
\end{equation}
\begin{equation}
    F^{DFP}(t) = F^{\mathcal{L}}(t) + w^{(\mathcal{R}\rightarrow\mathcal{L})}F^{(\mathcal{R}\rightarrow\mathcal{L})}(t),
\end{equation}
where the weight $w^{(\mathcal{R}->\mathcal{L})}$ is the pixel-wise cosine similarities between the warped stereo features $F^{(\mathcal{R}\rightarrow\mathcal{L})}(t)$ and the current left features $F^{\mathcal{L}}(t)$. The DFP enhancement features $F^{DFP}(t)$ are delivered to the next transformer decoder of QBS baseline instead of original monocular ones. We train BDFP via the original losses $\mathcal{L}_{baseline}$ represented as Eq.~(\ref{eq:qbsa}).

\textbf{Pseudo Stereo Generation.}
To exploit the stereo cues for even monocular surgical dataset, we applys a pseudo stereo generation mechanism to complement pseudo right images inspired by~\citep{watson2020learning}. As shown in the bottom row of Figure~\ref{fig:feature_propagation}, we apply an offline monocular depth network $\mathcal{G}$ to estimate the depth $\mathcal{Z}(t) = \mathcal{G}(\mathcal{I^{L}}(t))$ for left frame $\mathcal{I^{L}}(t)$. The disparities $\tilde{D}(t)$  are translated from estimated depth and scaled by a random scaling factor $d_{s}$ within a plausible range $[d_{\min}, d_{\max}]$ as repented in Eq.~(\ref{eq:disparities}). The scaling factor $d_{s}$ simulates different camera baseline and focal length. After that, we use a forward warping operation $\mathcal{F}$ to obtain every pixel of pseudo right view $\tilde{\mathcal{I}}^{\mathcal{R}}(t)$ from corresponding left view $\mathcal{I^{L}}(t)$. Eq.~(\ref{eq:pseduogeneration}) symbolizes this operation.
\begin{equation}
    \tilde{D}(t)= d_{s}\frac{\mathcal{Z}_{\max}(t)}{\mathcal{Z}(t)},
    \label{eq:disparities}
\end{equation}
\begin{equation}
    \tilde{\mathcal{I}}^{\mathcal{R}}(t) = \mathcal{F}(\tilde{D}(t), \mathcal{I^{L}}(t)).
    \label{eq:pseduogeneration}
\end{equation}
Different from previous methods~\citep{watson2020learning} that supplement stereo information only in training time, we make simulation not only in training stage with a randomly-sampled scale factor but also in inference stage with an average scale factor. A series of strategies are also applied to handle the unrealistic nature of generated image $\tilde{\mathcal{I}}^{\mathcal{R}}(t)$. To address `blurry' edges, a Sobel edge filter is used to identify the flying pixels and sharpen the disparity map. Instead of filling invisible regions with other images randomly sampled in \citep{watson2020learning}, we \textit{fill the missing regions with textures from a temporal frame or keep blank value with a valid mask} to relieve the effect of noise for the DFP module.

\subsection{Stereo-Temporal Set Classifier (STSCls)}
\begin{figure*}[!htbp]
	\centering
	\includegraphics[width=\textwidth]{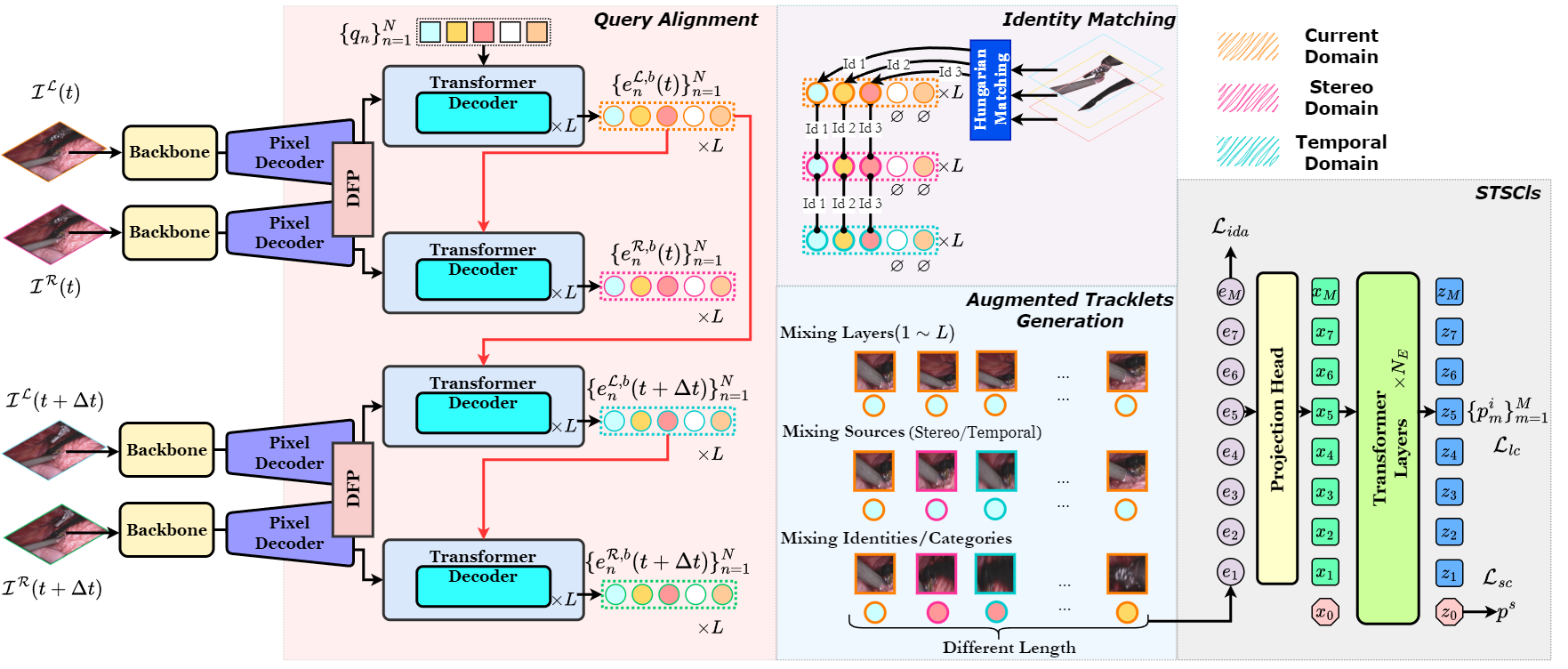}
	\caption{Schematic illustration of our STSCls module \& Training Pipeline.}
	\label{fig:setcls}
\end{figure*}
Forming tracklets from object queries is the first step for STSCls to make tracklet-level prediction. However, it is difficult for most surgical datasets to get video identity annotations, which describe instance correlations among frames. The naive tracklet sampling mechanism or training a track head is unrealizable for missing identity IDs. How to label pseudo identity ID for each object query by exploring temporal and stereo contexts is a major challenge for forming tracklets.

\label{sect:qa}
\textbf{How to generate pseudo identity ID for object queries?}

\underline{Query Alignment.} Considering that object queries are good representation for instances, they can make a bridge to align identities between frames. We present a simple but valid query alignment mechanism to align identity correlation across both temporal and stereo dimensions. As illustrate in ``Query Alignment" section of Figure ~\ref{fig:setcls}, given a pair of current frame $\mathcal{I^{L}}(t)$, stereo frame $\mathcal{I^{R}}(t)$, temporal frame $\mathcal{I^{L}}(t+\Delta{t})$ and stereo temporal frame $\mathcal{I^{R}}(t+\Delta{t})$, we replace the initial learnable query embeddings $\mathcal{Q}$ of BDFP for temporal $\mathcal{I^{L}}(t+\Delta{t})$ and stereo frame $\mathcal{I^{R}}(t)$ with object query embeddings of the current frame $\mathcal{I^{L}}(t)$. Then, we assume object queries with then same query index $n$ can correspond to same identity across \textbf{short-term} frames. From this, we can label pseudo identity ID of object query and sample tracklets based on query index $n$, which are easy to access.

\textbf{How to select object queries for tracklets generation?}

\underline{Identity Matching.} After query alignment, we get pseudo identity ID of all object queries. However, given a large amount of object queries not corresponding to any instances (non-object queries), identity matching is necessary to filter out the invalid queries before tracklets generation, which reduces uncontrollable training noise. Specifically, valid object queries of current frame $\mathcal{I^{L}}(t)$ are selected by Hungarian Matching with annotations as described in the ``Identity Matching" section of Figure ~\ref{fig:setcls}. For consecutive temporal $\mathcal{I^{L}}(t+\Delta{t})$ and stereo frames $\mathcal{I^{R}}(t)$, the corresponding ones with matched query indexes will also be maintained. The scheme can reduce training instability in some certain.

\textbf{How to augment tracklets for training STSCls?}

\underline{Augmented Tracklets Generation.} After identity matching, the number of valid object queries is far smaller than region proposals of RPN. Instead of sampling tracklets only based on pseudo identity IDs, various tracklet augmentation methods are applied to mitigate overfitting. Firstly, the intermediate object query embeddings from multi-layer Transformer decoder of BDFP have different perspectives to describe the same instance. From this, augmented tracklets can be produced by integrating intermediate object queries derived from various $L$ Transformer decoder layers. Owing to the generality, the augmentation can be extended to temporal and stereo frames. Secondly, integrating object queries from different sources, including temporal sequences and stereo pairs, enhances the diversity further. Thirdly, object queries with various identities or categories are mixed through a tracklet, significantly increasing the number of potential tracklets. Finally, for mitigating the long-tail problems in surgery duration, we sample multi-length tracklets, following the distribution inversely proportional to category portions. The eponymous section of Figure ~\ref{fig:setcls} shows the process and gives some augmented examples.

\textbf{STSCls Structure \textsc{\&} Formulation.}
STSCls aims to aggregating instance information contained in a tracklet and making a consolidated prediction to improve single frame results. As illustrated in ``STSCls" section of Figure~\ref{fig:setcls}, STSCls is composed of stacked $N_{E}$ Transformer layers like~\citep{hwang2022cannot}. Inputs for this module come from two sources: a trainable set classification token and instance tokens derived from a tracklet. Given a tracklet $s_{k} = \{e^{b}_{k,m}\}_{m=1}^{M}$ (where $M$ means tracklet length, k means identity ID), a projection head first encodes embedding items into instance tokens $\{x_m\}_{m=1}^{M}$. By inserting the set classification token $x_0$ and instance tokens $\{x_l\}_{l=1}^{L}$ into STSCls, the set classification token $x_0$ aggregates the overall contextual information via self-attention operation from instance tokens. The output embeddings $\{z_l\}_{m=0}^{M}$ of STSCls are further processed into class predictions including a set classification logit $p^{s}$ and local instance classification logits $\{p_m^{i}\}_{m=1}^M$. The former $p^{s}$ represents tracklet-level class prediction. We define $z_0$ as \textit{tracklet object query embedding} represented as $\mathbf{e^{s}}$ in the following.

\textbf{Training Losses.} For training \textbf{STSCls} with augmented tracklets, we design a joint training loss composed of global set classification loss $\mathcal{L}_{sc}$, local instance classification loss $\mathcal{L}_{lc}$ and identity alignment loss $\mathcal{L}_{ida}$. The global set classification loss $\mathcal{L}_{sc}$ is formulated as a typical cross entropy (\textbf{CE}) loss between set classification logit $p^{s}$ and tracklet-level ground truth  $y_{gt}^{s}$ as Eq.~(\ref{eq:sc_lc}). To accelerate the training process, the local instance classification loss $L_{lc}$ is applied to assist STSCls training with local instance classification logits $\{p_m^{i}\}_{m=1}^M$ and instance-level ground truths $\{y_{gt,m}^{i}\}_{m=1}^M$. The formulation is also a cross entropy loss as in Eq.~(\ref{eq:sc_lc}).

\begin{equation}
\label{eq:sc_lc}
    \mathcal{L}_{sc} = \mathbf{CE}(p^s, y_{gt}^{s}), \mathcal{L}_{lc} = \sum_{m=1}^M \mathbf{CE}(p_m^{i},y_{gt,m}^{i}).
\end{equation}
From the assumption that object queries with the same identity ID should be similar, the identity alignment loss $\mathcal{L}_{ida}$ lightly clusters the object query embeddings based on identities to assist queries alignment across stereo/temporal dimensions. In particular, we collect all $K$ valid object queries after identity matching in a batch and calculate similarities $\mathcal{M}^{sim}$ among them, where $m_{ij} \in \mathcal{M}^{sim}$ means the similarity of $i_{th}$ and $j_{th}$ object queries. The corresponding identity alignment ground-truths $\tilde{\mathcal{M}}$ measures whether any two object queries correspond to the same identity or not. The item $\tilde{m_{ij}} \in \tilde{\mathcal{M}}$ equals to 1 if $i_{th}$ and $j_{th}$ object queries share same identity ID and vice versa. The identity alignment loss Eq.~(\ref{eq:sim}) is set as a binary cross entropy~(\textbf{BCE}) loss. As represented in Eq.~(\ref{eq:STSCls}), the overall training loss $\mathcal{L}_{STSCls}$ is a combination of $\mathcal{L}_{sc}$, $\mathcal{L}_{lc}$, and $\mathcal{L}_{ida}$.

\begin{equation}
\label{eq:sim}
    \mathcal{L}_{ida} = \sum_{i=1}^K\sum_{j=1}^K \mathbf{BCE}(m_{ij}, \tilde{m_{ij}}).
\end{equation}
\begin{equation}
\label{eq:STSCls}
    \mathcal{L}_{STSCls} = \mathcal{L}_{sc} + \mathcal{L}_{lc} + \mathcal{L}_{ida}.
\end{equation}

\subsection{Location-Agnostic Classifier (LACls)}
LACls explicitly decouples mask classification with mask segmentation process to migrate the adverse influence of location biases. This module receives processed images that crop and mask original images based on mask predictions $\{m_{n}(t)\}_{n=1}^{N}$ of BDFP and outputs the location-agnostic class prediction $\{p_{n}^{a}(t)\}_{n=1}^{N}$. For simplicity, we use a vision Transformer pretrained on natural images as LACls in this work. We define the class token of the last layer as location-agnostic object query embedding $e^{a}$.

\begin{equation}
\label{eq:LACls}
\mathcal{L}_{LACls} = \sum_{n=1}^{\tilde{N}}\mathbf{CE}(p_{n}^a, \tilde{c_{n}}).
\end{equation}

Considering better trade-off between training costs and performance, we train this module only once for each fold offline. With the plug-and playable design, the trained module can be inserted into any instance segmentation architecture directly. For training this module offline, we use mask ground-truths instead of mask predictions and supervised by cross entropy loss $\mathcal{L}_{LACls}$ as in Eq.~(\ref{eq:LACls}). During inference, the module is cascaded directly after BDFP and makes classification based on mask segmentation predictions.

\begin{table*}[!ht] 
\begin{center}
    \caption{Performance of SOTA SIS methods on \evda and \evdb instrument segmentation datasets. (BF-Bipolar Forceps, PF-Prograsp Forceps, LND-Large Needle Driver, VS/SI- Vessel Sealer/ Suction Instrument, GR/CA- Grasping Retractor/Clip Applier, MCS-Monopolar Curved Scissors, UP-Ultrasound Probe)}
		\begin{tabular}{lrrrrrrrrrrrr}
			\hline
			\textbf{Method} &
			\textbf{Ch\_} &
			\textbf{ISI\_} &
			\multicolumn{7}{c}{\textbf{Instrument Classes IOU}} &
			\textbf{mc} \\ 
			\cline{4-10}
			\textbf{} &
			\textbf{\iou} &
			\textbf{\iou} &
			\textbf{BF} &
			\textbf{PF} &
			\textbf{LND} &
			\textbf{VS/ SI} &
			\textbf{GR/ CA} &
			\textbf{MCS} &
			\textbf{UP} &
			\textbf{\iou} \\ 
			\hline
			\rowcolor{tc} \multicolumn{13}{c}{\textbf{Dataset \evda}} \\
			\hline
			TernausNet-11~\citep{shvets2018automatic}     & 35.27 & 12.67 & 13.45 & 12.39 & 20.51 & 5.97   & 1.08   & 1.00  & 16.76  & 10.17 \\
			MF-TAPNET~\citep{jin2019incorporating}        & 37.35 & 13.49 & 16.39 & 14.11 & 19.01 & 8.11   & 0.31   & 4.09  & 13.40  & 10.77 \\
			ISINET~\citep{gonzalez2020isinet}      & 55.62 & 52.20 & 38.70 & 38.50 & 50.09 & 27.43  & 2.01   & 28.72 & 12.56  & 28.96 \\
			TraSeTR~\citep{zhao2022trasetr}        & 60.40 & 65.20 & 45.20 & 56.70 & 55.8  & 38.90  & 11.40  & 31.30 & 18.20  & 36.79 \\
			S3Net~\citep{baby2023forks}            & 72.54 & 71.99 & 75.08 & 54.32 & 61.84 & 35.50  & 27.47  & 43.23 & 28.38  & 46.55 \\
            MATIS (Frame)~\citep{ayobi2023matis}   & 62.74 & 68.79 & 66.18 & 50.99 & 52.23 & 32.84  & 15.71  & 19.27 & 23.90  & 37.30 \\
            MATIS (Full) ~\citep{ayobi2023matis}   & 71.36 & 66.28 & 68.37 & 53.26 & 53.55 & 31.89  & 27.34  & 21.34 & 26.53  & 41.09 \\
            QPD ~\citep{dhanakshirur2023learnable}  & 77.80 & 79.58 & 70.61 & 45.84 & 80.01 & 63.41 & \textbf{33.64}  & 66.57 & 35.28  & 49.92 \\
            \hline
            LACOSTE (S)              & 76.32 & 72.37 & 73.24 & 52.04 & 60.41 & 38.73  & 0.00   & 54.53 & 67.88  & 48.22 \\
            LACOSTE (B)               & 82.31 & 78.56 & \textbf{82.45} & 67.35 & 67.75 & \textbf{52.18}  & 15.53  & 74.33 & \textbf{81.87}  & 61.21 \\ 
            LACOSTE (L)                  & \textbf{83.41} & \textbf{80.35} & 80.39 & \textbf{68.93} & \textbf{80.84} & 42.26  & 13.16  & \textbf{90.02} & 80.96  & \textbf{63.73} \\ 
			\hline
			\rowcolor{tc} \multicolumn{13}{c}{\textbf{Dataset \evdb}} \\
			\hline
			TernausNet-11~\citep{shvets2018automatic} & 46.22 & 39.87 & 44.20 & 4.67  & 0.00     & 0.00      & 0.00      & 50.44  & 0.00     & 14.19 \\
			MF-TAPNET~\citep{jin2019incorporating}    & 67.87 & 39.14 & 69.23 & 6.10  & 11.68 & 14.00  & 0.91   & 70.24  & 0.57  & 24.68 \\
			ISINET~\citep{gonzalez2020isinet}         & 73.03 & 70.97 & 73.83 & 48.61 & 30.98 & 37.68  & 0.00   & 88.16  & 2.16  & 40.21 \\
			TraSeTR~\citep{zhao2022trasetr}           & 76.20 & \_    & 76.30 & 53.30 & 46.50 & 40.60  & 13.90  & 86.30  & 17.50 & 47.77 \\
			S3Net~\citep{baby2023forks}               & 75.81 & 74.02 & 77.22 & 50.87 & 19.83 & 50.59  & 0.00   & 92.12  & 7.44  & 42.58 \\
            MATIS (Frame)~\citep{ayobi2023matis}      & 82.37 & 77.01 & 83.35 & 38.82 & 40.19 & 64.49  & 4.32   & 93.18  & 16.17 & 48.65 \\
            MATIS (Full)~\citep{ayobi2023matis}       & 84.26 & 79.12 & 83.52 & 41.90 & 66.18 & 70.57  & 0.00   & 92.96  & \textbf{23.13} & 54.04 \\
            QPD ~\citep{dhanakshirur2023learnable}    & 77.77 & 78.43 & 82.80 & 60.94 & 19.96 & 49.70  & 0.00   & \textbf{93.93}  & 0.00 & 43.84 \\
            \hline
            LACOSTE (S)              & 85.20 & 82.41 & 85.21 & 70.75 & 68.02 & 62.64  & 12.81  & 91.98  & 0.00 & 55.92 \\
            LACOSTE (B)              & 86.48 & 85.09 & \textbf{85.93} & 75.68 & \textbf{77.56} & 72.99  & 29.63  & 92.56  & 15.48 & 64.26 \\ 
            LACOSTE (L)              & 86.78 & \textbf{85.68} & 85.71 & \textbf{77.68} & 67.97 & \textbf{76.39}  & \textbf{45.29}  & 93.27  & 22.27 & \textbf{66.94} \\ 
			\hline \\
		\end{tabular}
	\label{tab:comparison_EV17_18}
\vspace{-10mm}
\end{center}
\end{table*}

\begin{algorithm}[!htb]
\caption{LACOSTE Train Algorithm}
\label{alg:lacoste_train}
\setstretch{1.1}
\textbf{Input}: Batchsize $B$, Batch Item~(current frame $\mathcal{I^{L}}(t_i)$, stereo frame $\mathcal{I^{R}}(t_i)$, temporal frame $\mathcal{I^{L}}(t_i+\Delta t)$, temporal-stereo frame $\mathcal{I^{R}}(t_i+\Delta t)$), Pseudo Identity ID Label function $\mathcal{F}_{id}$.\\
\textbf{Params}: BDFP $\mathbf{\Phi_{B}}$, STSCls $\mathbf{\Phi_{S}}$, Initial Queries $\mathcal{Q}$.\\
\vspace*{-4mm}
\begin{algorithmic}[1]
\For {each iteration}
\Statex \quad\textbf{\textcolor{red}{Frame Step}}
\For {$i \leftarrow 1$ to $B$}
\For {$j \leftarrow $ in $\{t_i, t_i+\Delta{t}\}$}
\If{j = $t_i+\Delta{t}$}
\State $\mathcal{Q} = \{e_n^{b,\mathcal{L}}(t_i)\}_{n=1}^{N}$ \Comment{Query Alignment}
\Else
\State $\mathcal{Q} = \{q_n\}_{n=1}^{N}$
\EndIf
\State $\mathbf{\Phi_{B}}(\mathcal{I^{L}}(j)$,$\mathcal{I^{R}}(j))\xrightarrow{\mathcal{Q}}\bigcup\limits_{_\mathcal{L,R}}\{(e_n^{b}(j),p_n^{b}(j),m_n(j)\}_{n=1}^{N}$
\EndFor
\Statex \Comment{Label Pseudo Identity ID}
\State $\mathcal{F}_{id}(e_n^{b,{\mathcal{L}/\mathcal{R}}}(t_i/(t_i+\Delta t)))\rightarrow \tiny{(i-1)*N+n}$
\EndFor
\State Calculate $\mathcal{L}_{baseline}$
\Statex \quad\textbf{\textcolor{red}{Tracklet Step}}
\Statex \quad Filter out non-object queries \Comment{Identity Matching}
\Statex \quad  \Comment{Augmented Tracklets Generation}
\Statex \quad Mix multi-layers/sources/identities/categories queries
\Statex \quad Sample $N_s$ augmented tracklets
\For{$n \leftarrow 1$ to $N_s$}
\State $\mathbf{\Phi_{S}}(s_n) \rightarrow (e_n^s, c_n^s)$, $s_n$ means a tracklet
\EndFor
\State Calculate $\mathcal{L}_{STSCls}$
\Statex \textbf{Optimize $\mathbf{\Phi_{B}}$, $\mathbf{\Phi_{S}}$}
\EndFor
\end{algorithmic}
\end{algorithm}

\subsection{Overall Training \textsc{\&} Inference Pipeline.} In fact, LACOSTE can be trained in an end-to-end fashion. However, to reduce the training costs, we train the BDFP and STSCls jointly as Algorithm~\ref{alg:lacoste_train} and train LACls only once for each fold separately. In training stage, the formers are supervised with a combined loss as represented in Eq.~(\ref{eq:whole_loss}) and the latter is optimized with $\mathcal{L}_{LACls}$.
\begin{equation}
\label{eq:whole_loss}
    \mathcal{L}_{total} = \mathcal{L}_{baseline} + \mathcal{L}_{STSCls}.
\end{equation}
In inference stage, LACOSTE receives an 8-timestep (8$\times$2 frames) stereo clip centering at current frame $\mathcal{I^{L}}(t^*)$ and make end-to-end predictions as Algorithm~\ref{alg:lacoste_algo}. The final outputs of LACOSTE for current frame $\mathcal{I^{L}}(t^*)$ are composed of binary masks $\{m_{n}^{\mathcal{L}}(t^*)\}_{n=1}^{N}$ and class predictions $\{p_{n}^{f,\mathcal{L}}(t^*)\}_{n=1}^{N}$.
\section{Experiments}
\label{sec:experiment}
We experiment our LACOSTE on two public benchmark datasets of surgical videos, EndoVis2017 and EndoVis2018. Furthermore, dataset in other surgical domain, GraSP, is applied to validate the generality of method. Especially, we conduct comparison with state-of-the-art approaches, extensive ablation analysis on key components, and detailed visualization to validate the effectiveness of our approach.
\subsection{Datasets \textsc{\&} Implementation}
\textbf{EndoVis2017~(\evda).} The EndoVis2017 dataset~\citep{allan20192017} comprises ten video sequences captured from the da-Vinci robotic surgical system, accompanied by \textbf{instance} annotations for six distinct robotic instruments and an ultrasound probe. In order to facilitate equitable comparisons, we adopt a four-fold cross-validation in common with previous methods from a total of 1,800 frames (8 × 225). The fold-wise split yields 1,350 and 450 frames for training and validation, respectively.

\textbf{EndoVis2018~(\evdb).} The EndoVis2018 dataset~\citep{allan20202018} is collected from 2018 MICCAI Robotic Scene Segmentation Challenge. This dataset consists of 19 sequences, officially divided into 15 for training and 4 for testing. Each training sequence contains 149 frames recorded on a da Vinci X or Xi system during porcine training procedure. Each frame has a high resolution of 1280$ \times $1024. We use pre-defined training and validation splits from~\citep{shvets2018automatic} and annotate \textbf{instances} by ourselves for experiments.

\textbf{GraSP~(\texttt{GRASP}).} The GraSP dataset~\citep{ayobi2023towards,ayobi2024pixel} is a new curated benchmark that models surgical scene understanding. This dataset contains 13 sequences, officially divided into 8 for training and 5 for testing. This dataset provides multi-granularity including short-term segmentation and long-term recognition annotations. Each frame has a high resolution of 800$ \times $1280. It's noteworthy that GraSP provides monocular consecutive frames but only pixel-wise segmentation annotation for 3449 sparse frames. We use pre-defined training and validation splits from~\citep{ayobi2024pixel}.

\begin{gather}
    \mathrm{Ch}\_\iou = \frac{1}{K} \sum_{i=1}^{K} {\left(\frac{1}{\tilde{C_i}}\sum_{c=1}^{\tilde{C_i}} \frac{P_{i,c} \cap G_{i,c}}{P_{ic} \cup G_{i,c}}\right)} \notag \\
    \mathrm{ISI}\_\iou = \frac{1}{K} \sum_{i=1}^{K} {\left(\frac{1}{C}\sum_{c=1}^{C} \frac{P_{i,c} \cap G_{i,c}}{P_{ic} \cup G_{i,c}}\right)} \notag \\
    \mathrm{mc}\iou = \frac{1}{C} \sum_{c=1}^{C} {\left(\frac{1}{K}\sum_{i=1}^{K} \frac{P_{i,c} \cap G_{i,c}}{P_{i,c} \cup G_{i,c}}\right)} \label{eq:metrics}
\end{gather}

\textbf{Metrics.} For all datasets, we evaluate the performance on the Challenge IoU (Ch\_\iou) metric as proposed in the EndoVis2017 challenge and ISINet IoU (ISI\_\iou) and mean class IoU (mc\iou) metrics proposed in~\citep{gonzalez2020isinet}, to facilitate comparison. The formulas of metrics are represented in Eq.~(\ref{eq:metrics}), wherein $P$/$G$ mean segmentation predictions/ground truths, $C$/$\tilde{C_i}$ mean all classes/ground truth classes of current frame and $K$ represents number of frames.

\textbf{Implementation Details.}
Our framework is implemented based on Mask2Former and initialized by pretrained parameters with coco instance segmentation. All frames are resized into a range of (256, 1024) and crop $640 \!\times \! 512$ ($640 \!\times \! 400$ for \texttt{GRASP}) images randomly. The augmentation is consistent across the stereo, temporal, and current frames. We deploy the AdamW optimizer with a poly learning rate scheduler and use a base learning rate of $1e\!-\!4$. Batch size is set to 4 or 8 based on GPU size and the clip length is empirically set to 8. We select top five instances for each frame to merge semantic segmentation results like other methods e.g. \citep{baby2023forks,dhanakshirur2023learnable,ayobi2023matis}.

\begin{table*}[!ht]
\begin{center}
    \caption{Performance of SOTA SIS methods on \texttt{GRASP}. \textbf{Ch\_}\iou, \textbf{ISI\_}\iou and mc\iou are the same as above for semantic segmentation.}
        \begin{tabular}{lrrcrrr}
        \hline
        \textbf{Methods} &
        \multicolumn{2}{c}{\textbf{Instance Segmentation}} 
        & & \multicolumn{3}{c}{\textbf{Semantic Segmentation}} \\
        \cline{2-3}
        \cline{5-7}
        & $\mathrm{AP50_{box}}$ & $\mathrm{AP50_{segm}}$ & & \textbf{Ch\_}\iou & \textbf{ISI\_}\iou & \textbf{mc}\iou \\ 
        \hline
        \rowcolor{tc} \multicolumn{7}{c}{\textbf{Dataset \texttt{GRASP}}} \\
        \hline
        TernausNet-11~\citep{shvets2018automatic} & - & - & & 41.74 & 24.46 & 16.87 \\
        MF-TAPNet~\citep{jin2019incorporating}    & - & - & & 66.63 & 29.23 & 24.98 \\
        ISINet~\citep{gonzalez2020isinet}         & 79.85 & 78.29 & & 78.44 & 70.85 & 56.67 \\
        QPD~\citep{dhanakshirur2023learnable}     & 88.46 & 87.39 & & 83.89 & 82.56 & 74.36 \\
        TAPIS (Frame)~\citep{ayobi2024pixel}    & \textbf{92.65} & 91.71 & & 86.91 & 83.92 & 77.59 \\
        TAPIS (Full)~\citep{ayobi2024pixel}     & 91.72 & 90.34 & & \textbf{87.05} & 84.45 & 78.82 \\
        \hline
        LACOSTE (S)  & 89.69 & 90.83 & & 86.14 & 84.04 & 78.23 \\
        LACOSTE (B)  & 91.80 & \textbf{92.39} & & 86.77 & 83.95 & 79.44 \\
        LACOSTE (L)  & 90.72 & 92.15 & & 86.37 & \textbf{84.81} & \textbf{80.07} \\
        \hline
    \end{tabular}
    \label{tab:comparison_Grasp}
\vspace{-8mm}
\end{center}
\end{table*}

\subsection{Main Results}
\textbf{EndoVis 2017 \textsc{\&} 2018.} We compare our LACOSTE with state-of-the-art approaches including not only single frame methods, S3Net and QPD, but also temporal methods, MATIS. For fair comparison, we validate our approach based on Mask2Former with different backbones. For most experiments in this work, We implement our method based on \texttt{SwinBase} backbone. To compare with MATIS, we implement a parameters-equivalent one with the same \texttt{SwinSmall} backbone. QPD is implemented with MaskDino~\citep{li2023mask}, which is designed with a \texttt{SwinLarge} backbone and 300 queries. We also present a heavy-weight model with a \texttt{SwinLarge} backbone in this paper. For simplicity, we symbol \texttt{SwinBase}, \texttt{SwinSmall} and \texttt{SwinLarge} as \textbf{B}, \textbf{S} and \textbf{L} in the following parts. Results of different methods are presented in Table~\ref{tab:comparison_EV17_18}. We export the best results of comparison methods reported in their papers. For \evda, LACOSTE(\textbf{B}) and LACOSTE(\textbf{L}) outperforms all other methods while LACOSTE(\textbf{S}) is inferior to QPD because of less parameters. LACOSTE(\textbf{L}) improves over QPD by \textbf{7\%} Ch\_\iou,  \textbf{1\%} ISI\_\iou, and \textbf{28\%} mc\iou showing that temporal-stereo information improves the results by a considerable margin. Even for other temporal consistency methods, LACOSTE(\textbf{S}) outperforms MATIS by an improvement of \textbf{7\%} Ch\_\iou, \textbf{9\%} ISI\_\iou and \textbf{17\%} mc\iou while LACOSTE(\textbf{L}) improves \textbf{17\%} Ch\_\iou, \textbf{21\%} ISI\_\iou and \textbf{55\%} mc\iou respectively. For \evdb, LACOSTE is superior to single frame methods obviously. LACOSTE(\textbf{L}) improves over QPD by \textbf{12\%} Ch\_\iou, \textbf{9\%} ISI\_\iou, and \textbf{53\%} mc\iou. Moreover, LACOSTE(\textbf{S}) precedes MATIS slightly with \textbf{1\%} Ch\_\iou, \textbf{4\%} ISI\_\iou, and \textbf{3\%} mc\iou while LACOSTE(\textbf{L}) improves \textbf{3\%}, \textbf{8\%}, and \textbf{24\%}, respectively. As noted above, temporal information is more effective for \evdb and enhancing inter-instruments discrepancy is more important for \evda. The performance improvement between LACOSTE(\textbf{B}) and LACOSTE(\textbf{S}) is more obvious than that changing the backbone from \texttt{SwinBase} to \texttt{SwinLarge}. For better trade-off between training costs and performance, we validate the following experiments and analysis based on LACOSTE(\textbf{B}).

\begin{table*}[!ht]
	\begin{center}
    \caption{Performance of key components on \evda and \evdb. \textcolor{blue}{\checkmark} in LACls column means fine-tuning LACls.}
		\begin{tabular}{cccrrrrrrrrrrrr}
			\hline
			\multicolumn{3}{c}{\textbf{Key Componets}} & \textbf{Ch\_} & \textbf{ISI\_} &
			\multicolumn{7}{c}{\textbf{Instrument Classes IOU}} &
			\textbf{mc} \\
			\cline{6-12}
			DFP & STSCls & LACls&
			\textbf{\iou} &
			\textbf{\iou} &
			\textbf{BF} &
			\textbf{PF} &
			\textbf{LND} &
			\textbf{VS/ SI} &
			\textbf{GR/ CA} &
			\textbf{MCS} &
			\textbf{UP} &
			\textbf{\iou} \\ 
			\hline
			\rowcolor{tc} \multicolumn{15}{c}{\textbf{Dataset \evda}} \\
			\hline
			 & & & 75.12 & 71.68 & 60.42 & 62.97 & 60.88 & 36.29   & 3.14   & 30.22  & 46.00  & 44.48 \\
			 \checkmark & & & 78.34 & 74.13 & 74.19 & 64.57 & 55.17 & 38.01   & 10.19  & 35.22  & 49.24  & 47.88 \\
			 \checkmark & T & & 81.21 & 78.02 & 74.72 & \textbf{68.63} & \textbf{90.03} & \textbf{53.49}   & \textbf{27.52}  & 41.49  & 52.62  & 55.32 \\
			\checkmark & S & & 81.12 & 78.19 & 74.65 & 62.13 & 66.80 & 37.66   & 14.26  & 77.51  &53.64 & 55.50 \\
			\checkmark & ST & & 81.89 & 78.52 & 82.77 & 67.86 & 67.90 & 51.70   & 15.40  & 73.90  & 81.87  & 61.26 \\
             \checkmark & ST & \checkmark & 82.31 & 78.56 & 82.45 & 67.35 & 67.75 & 52.18   & 15.53  & 74.33  & 81.87  & 61.21 \\
             \checkmark & ST & \textcolor{blue}{\checkmark} & \textbf{82.90} & \textbf{79.05} & \textbf{82.89} & 67.24 & 68.02 & 52.71   & 17.53  & \textbf{74.59}  & \textbf{82.09}  & \textbf{61.76} \\
			\hline
			\rowcolor{tc} \multicolumn{15}{c}{\textbf{Dataset \evdb}} \\
			\hline
			 & &    & 82.93 & 80.54 & 85.23 & 69.42  & 45.37 & 56.36  & 0.00      & 91.99  & 0.00     & 49.77 \\
			 \checkmark & &   & 84.33 & 82.92 & 85.42 & 74.01  & 64.12 & 53.39  & 0.00      & 92.74  & 3.38  & 53.29 \\
			 \checkmark & T & & 85.72 & 84.08 & \textbf{86.37} & 71.77 & 67.26 & \textbf{76.20} & 6.71  & 92.42 & \textbf{21.99} & 60.39 \\
			 \checkmark & S & & 85.90 & 83.99 & 85.90 & 72.28 & 72.36 & 71.51 & 0.00 & \textbf{93.18} & 0.00 & 56.46 \\
			 \checkmark & ST & & 86.02 & 84.42 & 85.94 & \textbf{75.99} & \textbf{79.25} & 70.37 & 2.53 & 92.78 & 7.44 & 58.12 \\
             \checkmark & ST & \checkmark & 86.48 & 85.09 & 85.93 & 75.68 & 77.56 & 72.99  & 29.63  & 92.56 & 15.48 & 64.26 \\ 
             \checkmark & ST & \textcolor{blue}{\checkmark}& \textbf{86.66} & \textbf{85.35} & 85.89 & 75.37 & 75.94 & 73.26 & \textbf{53.58} & 92.77 & 12.28 & \textbf{67.01} \\ 
			\hline \\
		\end{tabular}
	\label{tab:performance_key_componets}
 \end{center}
 \vspace{-7mm}
\end{table*}

\textbf{GraSP.} We compare our LACOSTE with TAPIS~\citep{ayobi2024pixel} which is a multi-task method for surgical scene understanding and other state-of-the-art approaches reported in their paper. Evaluations are implemented not only in semantic segmentation but also in instance segmentation. In particular,  Ch\_\iou, ISI\_\iou and mc\iou are selected for semantic performance while $\mathrm{AP50_{box}}$ and $\mathrm{AP50_{segm}}$ are used for instance performance. Results of different methods are presented in Table~\ref{tab:comparison_Grasp}. Observe that LACOSTE(\textbf{L}) attains the best overall results of 80.07 mc\iou and 84.81 ISI\_\iou. In the task of instance segmentation, as measured by the $\mathrm{AP50_{segm}}$ metric, LACOSTE (\textbf{B}) demonstrates superior performance compared to all other evaluated methods, while LACOSTE (\textbf{L}) achieves the second highest performance. TAPIS (Frame) and TAPIS (Full) respectively outperform LACOSTE in $\mathrm{AP50_{box}}$ and Ch\_\iou. This superior performance is likely attributable to the multi-task architecture and the incorporation of additional annotations, such as those for detection and phase recognition. But promisingly, LACOSTE achieves peak or second segmentation performance across most evaluation metrics. This result suggests that our method is robust and generality to various surgical domain. When comparing the improvement of LACOSTE with that of \evda and \evdb, the enhancement may be constrained by limited binary mask segmentation for \texttt{GRASP}, which is not the primary design focus of this paper.

\subsection{Ablation Study}
We conduct ablation experiments to validate the effectiveness of different key components in the proposed method and obtain seven configurations:
\begin{enumerate}[(a)]
    \item \textbf{Baseline:} We train the pure Mask2former network as the baseline;
    \item \textbf{Baseline~(DFP):} We train the baseline Mas2former with DFP;
    \item \textbf{Baseline~(DFP,~T):} We train STSCls with augmented tracklets by mixing object queries across only temporal dimension;
    \item \textbf{Baseline~(DFP,~S):} We train STSCls with augmented tracklets by mixing object queries across only stereo dimension;
    \item \textbf{Baseline~(DFP,~ST):} We train STSCls with augmented tracklets by mixing object queries across both temporal and stereo dimensions;
    \item \textbf{Baseline~(DFP,~ST,~LACls):} We apply a LACls without fine-tuning after (e);
    \item \textbf{Baseline~(DFP,~ST,~\textcolor{blue}{LACls}):} We fine-tune a LACls after (e) for some iterations;
\end{enumerate}

\textbf{Effectiveness of Key Components.} The results on \evda and \evdb are presented in Table~\ref{tab:performance_key_componets}. We observe that our baseline based on \texttt{SwinBase} backbone obtains reasonable results with Ch\_\iou and ISI\_\iou over 70 on all tasks of both datasets. Purely introducing DFP module on stereo dimension, (b) achieves better results by \textbf{4\%} Ch\_\iou, \textbf{3\%} ISI\_\iou, \textbf{8\%} mc\iou for \evda and \textbf{2\%} Ch\_\iou, \textbf{3\%} ISI\_\iou, \textbf{7\%} mc\iou for \evdb. Incorporating temporal-stereo consistency cues by STSCls, (c) (d) (e) further improves the segmentation performance in all evaluation metrics of both datasets. For different augmented tracklets configure settings, we observe that (e) is superior to the other two on most metrics especially mcIoU. We compare (e) with (b) to evaluate the effectiveness of STSCls intuitively. Ch\_\iou and ISI\_\iou on \evda are increased respectively \textbf{3.55}, and \textbf{4.39}. The same trends rising \textbf{1.69} Ch\_\iou and \textbf{1.5} ISI\_\iou can also be observed on \evdb. Adding our LACls without any fine-tuning operations, our full model continues boosting the results with \textbf{0.4} Ch\_\iou and \textbf{0.4$\sim$1} ISI\_\iou gain, peaking at \textbf{82.31} Ch\_\iou, \textbf{78.56} ISI\_\iou on \evda and \textbf{86.48} Ch\_\iou \textbf{85.09} ISI\_\iou on \evdb. The architecture achieves \textbf{82.90} Ch\_\iou, \textbf{79.05} ISI\_\iou, \textbf{61.76} mc\iou for \evda and \textbf{86.66} Ch\_\iou, \textbf{85.35} ISI\_\iou, \textbf{67.01} mc\iou for \evdb with some iterations fine-tuning. To reduce training costs, we report our methods based on \texttt{SwinSmall}, \texttt{SwinBase} and \texttt{SwinLarge} backbone in last subsection without any fine-tuning.

\begin{table*}[!ht]
\begin{center}
    \caption{Performance of ensemble items of LACOSTE on \evda \& \evdb.}
		\begin{tabular}{lrrrrrrrrrrrr}
			\hline
			\textbf{Method} &\textbf{Ch\_} & \textbf{ISI\_} &
			\multicolumn{7}{c}{\textbf{Instrument Classes IOU}} &
			\textbf{mc} \\ 
			\cline{4-10}
			\textbf{} &
			\textbf{\iou} &
			\textbf{\iou} &
			\textbf{BF} &
			\textbf{PF} &
			\textbf{LND} &
			\textbf{VS/ SI} &
			\textbf{GR/ CA} &
			\textbf{MCS} &
			\textbf{UP} &
			\textbf{\iou} \\ 
			\hline
			\rowcolor{tc} \multicolumn{13}{c}{\textbf{Dataset \evda}} \\
			\hline
            LACOSTE~(F)  & 81.39 & 76.18 & 80.19 & 60.69 & 63.80 & 38.08   & 10.93  & 37.05  & 40.62  & 49.68 \\
            LACOSTE~(S)  & 81.60 & 78.25 & 81.64 & \textbf{67.97} & 67.60 & 51.69   & \textbf{16.15}  & 48.70  & 81.62  & 59.05 \\
            LACOSTE~(A)  & 81.76 & 76.43 & 79.13 & 59.74 & 62.74 & 39.14   & 11.68  & 52.76  & 40.52  & 49.38 \\
			LACOSTE      & \textbf{82.31} & \textbf{78.56} & \textbf{82.45} & 67.35 & \textbf{67.75} & \textbf{52.18}   & 15.53  & \textbf{74.33}  & \textbf{81.87}  & \textbf{61.21} \\
			\hline
			\rowcolor{tc} \multicolumn{13}{c}{\textbf{Dataset \evdb}} \\
			\hline
            LACOSTE~(F)  & 85.34 & 83.73 & 85.58 & 74.63 & 72.15 & 64.94  & 13.68  & 91.75 & 5.62 & 58.34 \\
            LACOSTE~(S)  & 85.85 & 84.36 & \textbf{86.09} & \textbf{76.42} & 77.39 & 67.12  & 8.03  & \textbf{92.77} & 0 & 58.26 \\
            LACOSTE~(A)  & 86.04 & 84.59 & 85.87 & 75.00 & 74.05 & 71.58  & 17.89  & 92.36 & \textbf{17.40} & 62.02 \\
            LACOSTE      & \textbf{86.48} & \textbf{85.09} & 85.93 & 75.68 & \textbf{77.56} & \textbf{72.99}  & \textbf{29.63}  & 92.56 & 15.48 & \textbf{64.26} \\
			 \hline \\
		\end{tabular}
	\label{tab:eachpart}
 \end{center}
 \vspace{-8mm}
\end{table*}

\textbf{Performance of Ensemble Items.} We list the performance of each item in final ensemble results as Table~\ref{tab:eachpart}. LACOSTE~(F), LACOSTE~(S), LACOSTE~(A) mean the results of the Frame, Tracklet, and L-Agnostic step, respectively. The effectiveness of ensemble mechanism can be validated on both datasets. Among all items, STSCls of Tracklet step exerts the most significant influence on ISI\_\iou and mc\iou while LACls of L-Agnostic step demonstrates superior performance in Ch\_\iou. In the context of instrument classes IOU, items exhibit mutual complementarity across various instrument types.

\begin{table*}[!ht]
	\begin{center}
    \caption{Performance of pseudo stereo mechanism on \evda \& \evdb.}
		\begin{tabular}{lrrrrrrrrrrrr}
			\hline
			\textbf{Method} &\textbf{Ch\_} & \textbf{ISI\_} &
			\multicolumn{7}{c}{\textbf{Instrument Classes IOU}} &
			\textbf{mc} \\ 
			\cline{4-10}
			\textbf{} &
			\textbf{\iou} &
			\textbf{\iou} &
			\textbf{BF} &
			\textbf{PF} &
			\textbf{LND} &
			\textbf{VS/ SI} &
			\textbf{GR/ CA} &
			\textbf{MCS} &
			\textbf{UP} &
			\textbf{\iou} \\ 
			\hline
			\rowcolor{tc} \multicolumn{13}{c}{\textbf{Dataset \evda}} \\
			\hline
			LACOSTE  & 82.31 & 78.56 & 82.45 & 67.35 & 67.75 & 52.18   & 15.53  & 74.33  & 81.87  & 61.21 \\
            LACOSTE~(P)  & 82.10 & 78.00 & 78.36 & 67.24 & 65.75 & 39.43   & 27.97  & 44.08  & 75.95  & 52.19 \\
			\hline
			\rowcolor{tc} \multicolumn{13}{c}{\textbf{Dataset \evdb}} \\
			\hline
            LACOSTE  & 86.48 & 85.09 & 85.93 & 75.68 & 77.56 & 72.99  & 29.63  & 92.56 & 15.48 & 64.26 \\
            LACOSTE~(P)  & 85.57 & 84.20 & 86.46 & 77.43 & 73.70 & 54.09  & 2.66   & 93.69 & 0.00  & 55.43 \\
			 \hline \\
		\end{tabular}
	\label{tab:pseudo_stereo}
 \end{center}
 \vspace{-8mm}
\end{table*}

\begin{table*}[!ht]
\begin{center}
    \caption{Influence of non-object masking mechanism on \evda and \evdb.}
		\begin{tabular}{lrrrrrrrrrrrr}
			\hline
			\textbf{Method} &\textbf{Ch\_} & \textbf{ISI\_} &
			\multicolumn{7}{c}{\textbf{Instrument Classes IOU}} &
			\textbf{mc} \\ 
			\cline{4-11}
			\textbf{} &
			\textbf{\iou} &
			\textbf{\iou} &
			\textbf{BF} &
			\textbf{PF} &
			\textbf{LND} &
			\textbf{VS/ SI} &
			\textbf{GR/ CA} &
			\textbf{MCS} &
			\textbf{UP} &
			\textbf{\iou} \\ 
			\hline
			\rowcolor{tc} \multicolumn{13}{c}{\textbf{Dataset \evda}} \\
			\hline
            LACOSTE  & 82.31 & 78.56 & 82.45 & 67.35 & 67.75 & 52.18 & 15.53  & 74.33  & 81.87  & 61.21 \\
            LACOSTE~(M)  & 81.53 & 77.57 & 81.03 & 61.07 & 63.57 & 40.28   & 14.26  & 78.52  & 40.83  & 52.66 \\
			\hline
			\rowcolor{tc} \multicolumn{13}{c}{\textbf{Dataset \evdb}} \\
			\hline
            LACOSTE  & 86.09 & 84.50 & 86.10 & 76.00 & 78.40 & 71.05  & 7.51   & 92.56  & 15.48 & 58.80 \\
            LACOSTE~(M)  & 86.48 & 85.09 & 85.93 & 75.68 & 77.56 & 72.99  & 29.63   & 92.56  & 15.48 & 64.26 \\ 
			\hline \\
		\end{tabular}
	\label{tab:non_object}
 \end{center}
 \vspace{-8mm}
\end{table*}

\textbf{Pseudo Stereo Generation.} To validate the effectiveness of pseudo stereo generation mechanism for only monocular dataset, we experment LACOSTE respectively with real right images and generated pseudo right images for both datasets. The results are presented in Table~\ref{tab:pseudo_stereo}, where LACOSTE(P) means the experiments with pseudo generation. We observe that the results decrease \textbf{0.21} in Ch\_\iou, \textbf{0.56} in ISI\_\iou, and \textbf{9} in mc\iou but outperform the baseline (a) on \evda. The results on \evdb decrease \textbf{0.91} in Ch\_\iou, \textbf{0.89} in ISI\_\iou, and \textbf{9} in mc\iou. The decreasing performance is within an acceptable range. The slight misclassification on non-existent classes of validation dataset makes the mc\iou metric decrease obviously. The pseudo stereo generation mechanism can complement monocular dataset to some extent.

\textbf{Non-object Masking.} For surgical domain, the same instance can move out of view through a stereo clip, where object query fails to match with any instance in some frames. As mentioned above, object query not coresponding to any instance is named as \textit{non-object query}, which can be determined based on BDFP classification. We also investigate the influence of masking out non-objects queries within a self-attention operation of STSCls in inference stage. The results are presented in Table~\ref{tab:non_object}, where LACOSTE(M) means the experiments with masking operation. We find that masking non-object queries is superior for \evdb but inferior for \evda. The rationale behind this may be that \evdb contains more move-out-of-view situations than \evda dataset. Masking operation decreases the influence of noise information.

\begin{table}[!ht]
\begin{center}
   \small
    \caption{Inference time for proposed components of LACOSTE on \evda.}
		\begin{tabular}{ccccrrr}
            \hline
			\multicolumn{3}{c}{\textbf{Key Componets}} & Inference & \textbf{Ch}\_ & \textbf{ISI}\_ & \textbf{mc}\\
            DFP & STSCls & LACls & Time~(s) & \iou & \iou & \iou \\
			\hline
			& & & 0.219 & 75.12 & 71.68 & 44.48\\
            \checkmark & & & 0.323 & 78.34 & 74.13 & 47.88\\
            \checkmark & ST & & 0.343 & 81.89 & 78.52 & 61.26\\
            \checkmark & ST & \checkmark & 0.371 & 82.31 & 78.56 & 61.21\\
            \hline
		\end{tabular}
	\label{tab:inferencetime_mod}
\end{center}
\end{table}

\textbf{Inference Time.} We analyze the inference time of each proposed component as Table~\ref{tab:inferencetime_mod}. For improving model efficiency and reducing inference time, we introduce a memory bank to record intermediate results including features, embeddings and predictions which can be reused across different clips. Furthermore, we also record corresponding valid object queries indices which can reduce the computations of LACls. In comparison with baseline model, LACOSTE requires additional computation for temporal and stereo contexts, but the performance gain over additional latency is substantial. We expect that inference optimization methods (e.g., TensorRT) can further reduce the latency of model.

\textbf{Other Metrics.} We also provide some other metrics to evaluate our methods from different perspectives. In particular, we select Dice similarity coefficient (DSC)/F1 score, mean class Dice (mcD), Hausdorff distance (HD) and average surface distance (ASD). The former two metrics are commonly used to evaluate the effectiveness of segmentation methods. HD and ASD are sensitive to the boundaries of segments, which represent segmentation completeness in some certain. The results are shown in Table~\ref{tab:other_metrics}, where baseline means the results of original Mask2Former. For all benchmarks, LACOSTE(L) demonstrates superior performance in both DSC and mcD metrics. In the assessment of segmentation completeness, LACOSTE(L) demonstrates inferior performance compared to LACOSTE(B) as indicated by higher values in HD and ASD metrics for \evda and \evdb while LACOSTE(L) exhibits superiority in \texttt{GRASP}.

\begin{table*}[!ht]
\small
\begin{center}
    \caption{More metrics of our methods in three surgical instrument datasets. \textbf{DSC($\uparrow$)} and \textbf{mcD($\uparrow$)} can show the segmentation accuracy. \textbf{HD($\downarrow$)} and \textbf{ASD($\downarrow$)} show the segmentation completeness.}
		\begin{tabular}{l|rrrr|rrrr|rrrr}
            \hline
			\textbf{Method} & \multicolumn{4}{c|}{\textbf{\evda}} & \multicolumn{4}{c|}{\textbf{\evdb}} & \multicolumn{4}{c}{\texttt{GRASP}}\\
			\cline{2-13}
			\textbf{} &
			\multicolumn{1}{c}{\textbf{DSC($\uparrow$)}} &
			\multicolumn{1}{c}{\textbf{mcD($\uparrow$)}} &
            \multicolumn{1}{c}{\textbf{HD($\downarrow$)}} &
            \multicolumn{1}{c|}{\textbf{ASD($\downarrow$)}} &
			\multicolumn{1}{c}{\textbf{DSC($\uparrow$)}} &
			\multicolumn{1}{c}{\textbf{mcD($\uparrow$)}} &
            \multicolumn{1}{c}{\textbf{HD($\downarrow$)}} &
            \multicolumn{1}{c|}{\textbf{ASD($\downarrow$)}} &
			\multicolumn{1}{c}{\textbf{DSC($\uparrow$)}} &
			\multicolumn{1}{c}{\textbf{mcD($\uparrow$)}} &
            \multicolumn{1}{c}{\textbf{HD($\downarrow$)}} &
            \multicolumn{1}{c}{\textbf{ASD($\downarrow$)}} \\
			\hline
			Baseline    & 75.49 & 47.15 & 163.60 & 39.93 & 84.38 & 52.32 & 108.79 & 13.90 & 84.85 & 76.97 & 106.04 & 13.57 \\
			LACOSTE(S)  & 76.46 & 51.11 & 134.76 & 19.68 & 86.33 & 59.02 & 142.04 & 18.62 & 88.06 & 82.68 & 95.71 & 18.62 \\
            LACOSTE(B)  & 82.22 & 64.20 &  \textbf{90.26} & 14.89 & 88.94 & 64.26 & \textbf{103.90} & \textbf{12.82} & 87.97 & 83.89 & 95.80 & 10.21 \\
            LACOSTE(L)  & \textbf{84.11} & \textbf{66.91} &  94.54 & \textbf{12.60} & \textbf{89.68} & \textbf{70.90} & 122.45 & 15.80 & \textbf{88.72} & \textbf{84.39} & \textbf{94.35} & \textbf{9.89} \\
            \hline
		\end{tabular}
	\label{tab:other_metrics}
\end{center}
\end{table*}

\begin{figure*}[!ht]
	\centering
	\includegraphics[width=\textwidth]{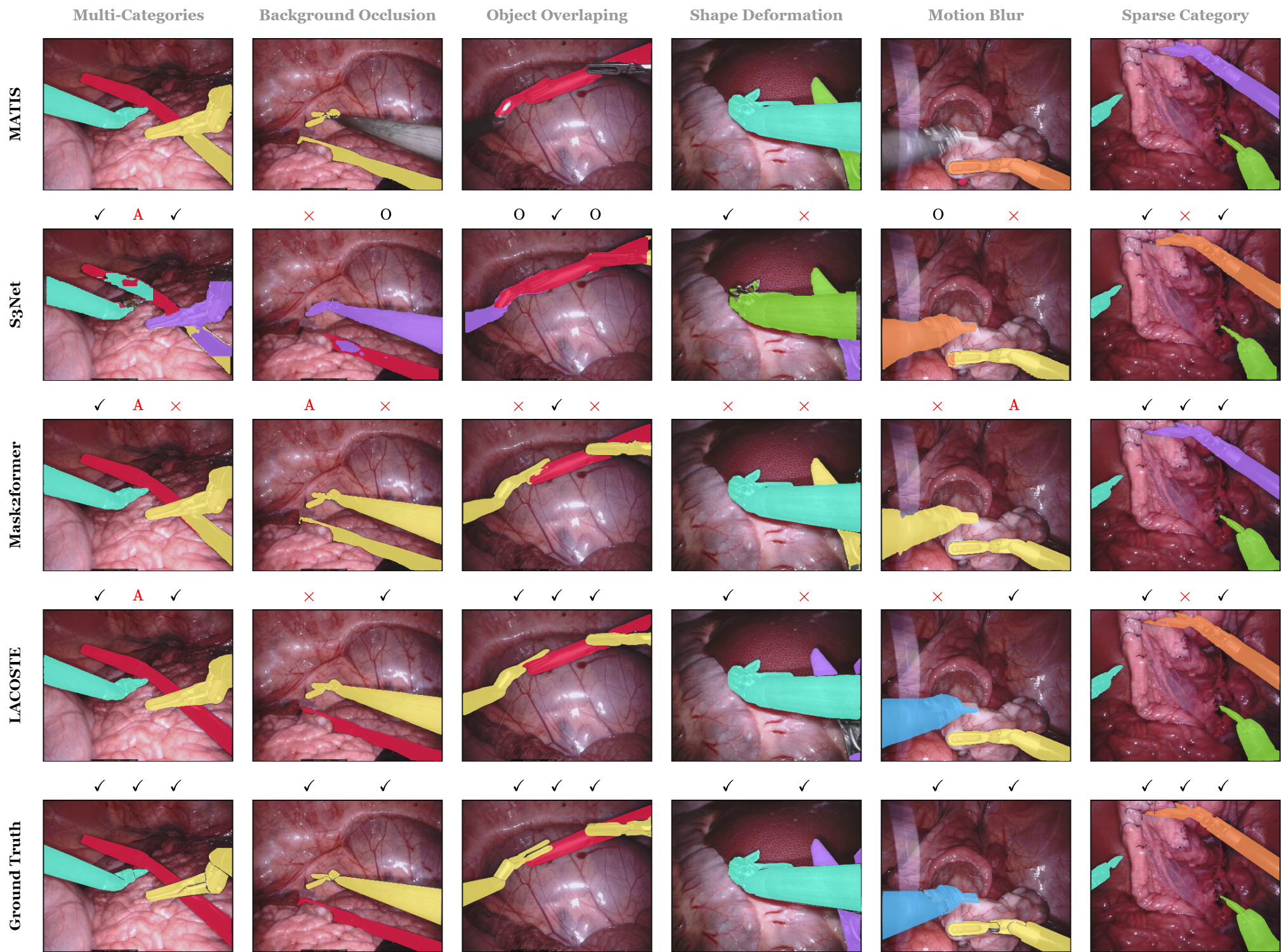}
	\caption{Qualitative analysis for SIS results: different symbols are used to show the results; $\checkmark$ represents the instance labeled correctly, \textcolor{red}{$\times$} shows the misclassified instance, and ‘O’ represents missed instance, The letter \textcolor{red}{A} is employed to denote instances characterized by ambiguity, wherein the selection of the appropriate instrument class is rendered uncertain. This ambiguity may arise from factors such as over-segmentation or the presence of multiple instances of instrument classes within the same spatial region.} 
	\label{fig:segment_results}
\end{figure*}

\subsection{Qualitative Analysis}
\begin{figure*}[!ht]
	\centering
	\includegraphics[width=\textwidth]{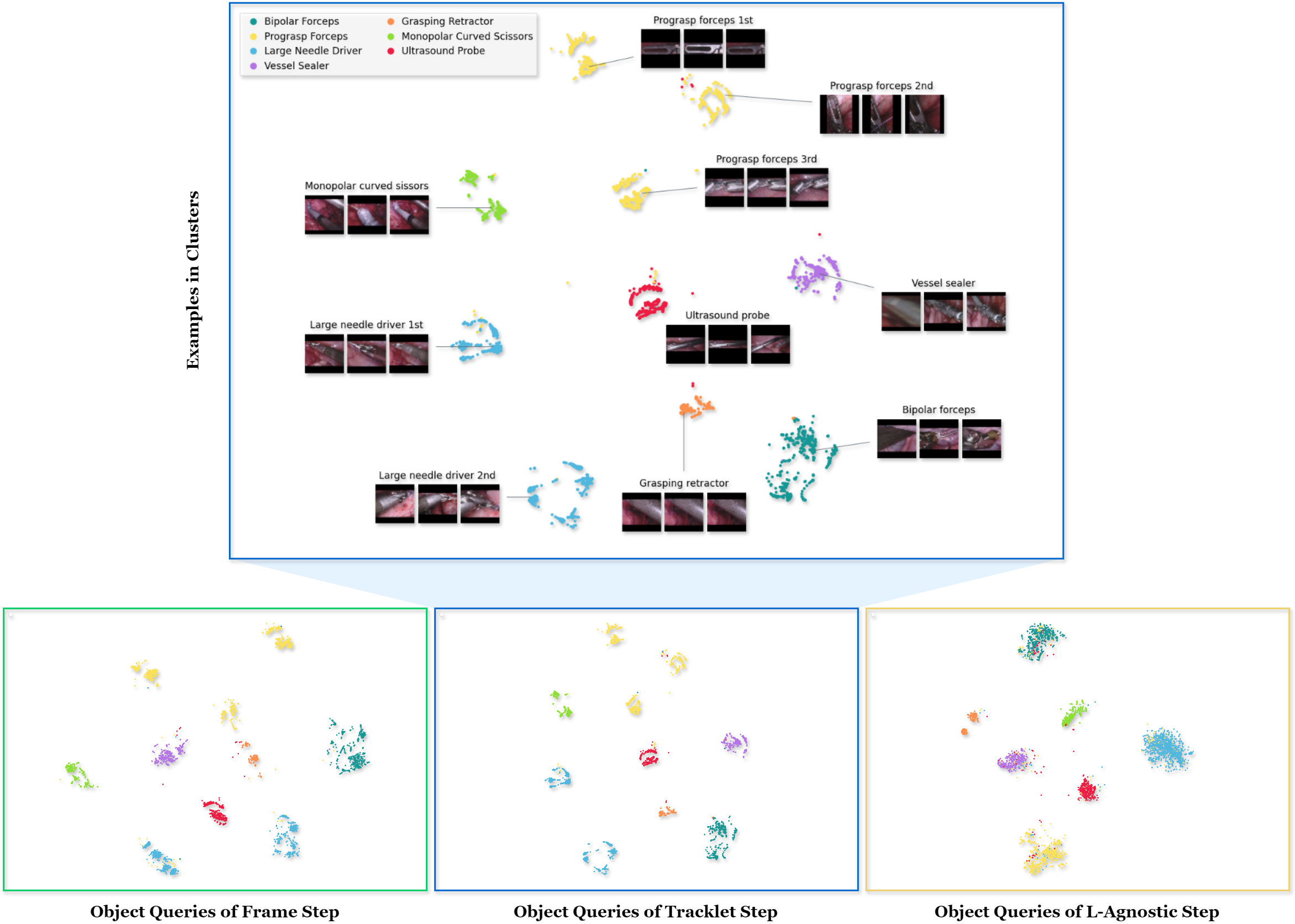}
	\caption{T-SNE analysis of query embedding space from each step. Instance examples of each cluster in tracklet object queries space are also visualized.}
	\label{fig:embeddingspace}
	\vspace{-5mm}
\end{figure*}

\textbf{Segmentation Results.} Figure~\ref{fig:segment_results} illustrates the qualitative segmentation results. We show the comparative results of a single frame approach named S3Net, a typical temporal consistency approach named MATIS, our baseline, and our proposed approach. The segmentation results are distributed across different situations including multi-categories, background occlusion, object overlapping, shape deformation, motion blur, and sparse category. We show better segmentation in most situations. LACOSTE also focuses on perceiving the temporal context and performs well in classifying motion blur, background occlusion and shape deformation circumstances. 

\textbf{Query Embedding Space Analysis.} We make t-SNE analysis of frame object queries embeddings $e^b$ from BDFP (\textbf{FOE}), tracklet object queries embeddings $e^s$ from STSCls (\textbf{TOE}) and location-agnostic object queries embeddings $e^a$ from LACls (\textbf{LAOE}) as shown in Figure~\ref{fig:embeddingspace}. For both FOE and TOE, query embedding spaces are partitioned into discriminative clusters and the latter exhibits a more compact structure. Additionally, except the Large Needle Driver (LND) and Prograsp Forceps (PF), the queries embeddings are distributed around one cluster. In Figure~\ref{fig:embeddingspace}, we visualize some instance examples for every cluster in TOE. As illustrated in `Large Needle Driver 1st' and `Large Needle Driver 2nd' instance examples, the location and orientation of the category have a significant change which arises the cluster deviation. Furthermore, rather than `Large Needle Driver 2nd', the cluster of `Large Needle Driver 1st' is closer to `Monopolar Curved Scissors'. The location bias may have a negative influence in semantic classification. After inducing LACls, the multi-clusters of LND and PF in LAOE space can collapse into only one. It is worth noting that the embedding projection points of TOE space are fewer than the other two because the examples belonging to one clip project into the same location.

\begin{figure*}[!htbp]
	\centering
	\includegraphics[width=0.9\textwidth]{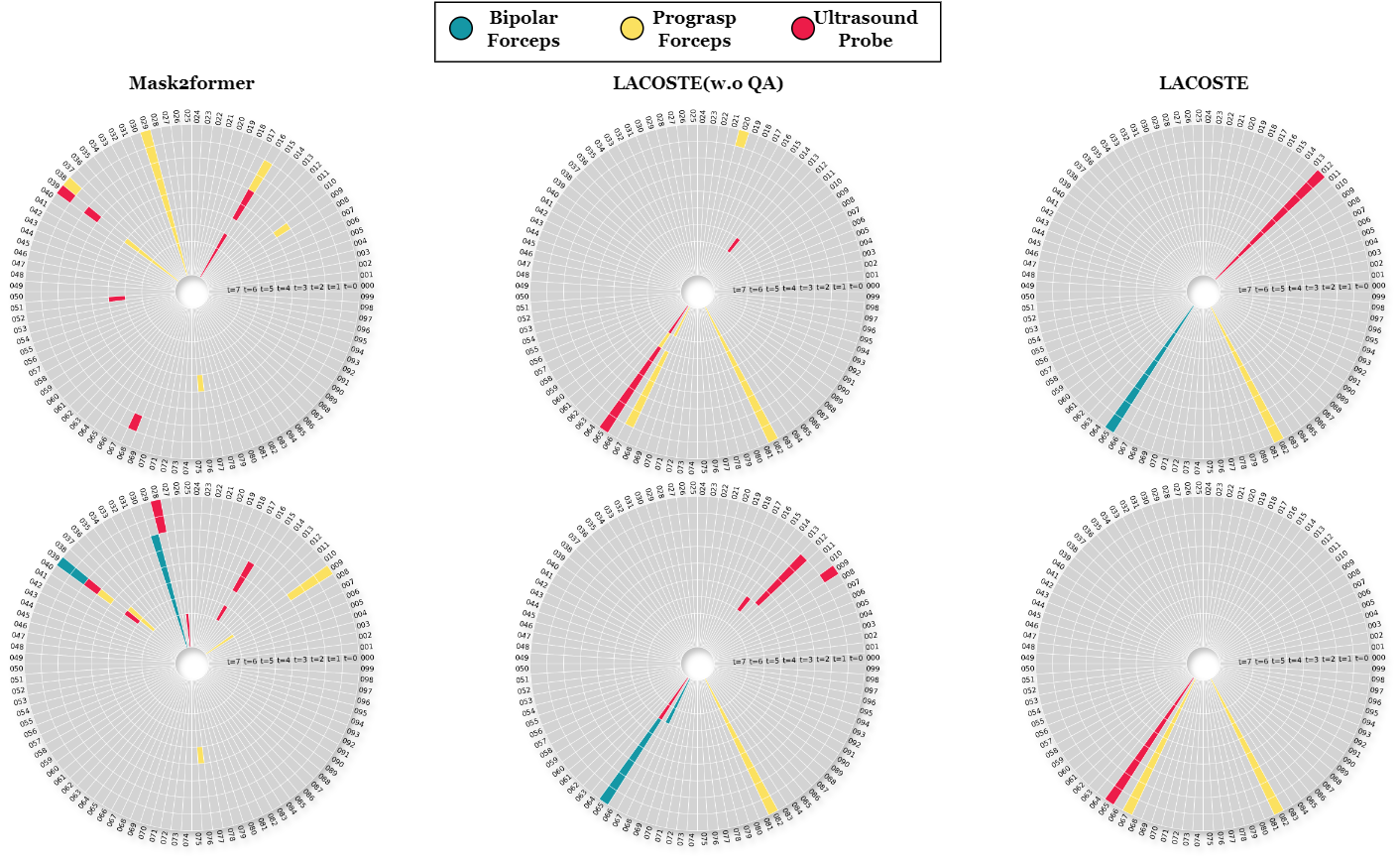}
	\caption{Comparison of Query Alignment Situations. The rings from outermost to innermost represent eight temporal frames. The counterclockwise direction of each ring represents 100 query indexes. LACOSTE(w.o QA) means our proposed method with only identity alignment loss constrains while LACOSTE means the full model.}
	\label{fig:sequecnce_identity}
 \vspace{-3mm}
\end{figure*}

\begin{figure*}[!htbp]
	\centering
	\includegraphics[width=0.8\textwidth]{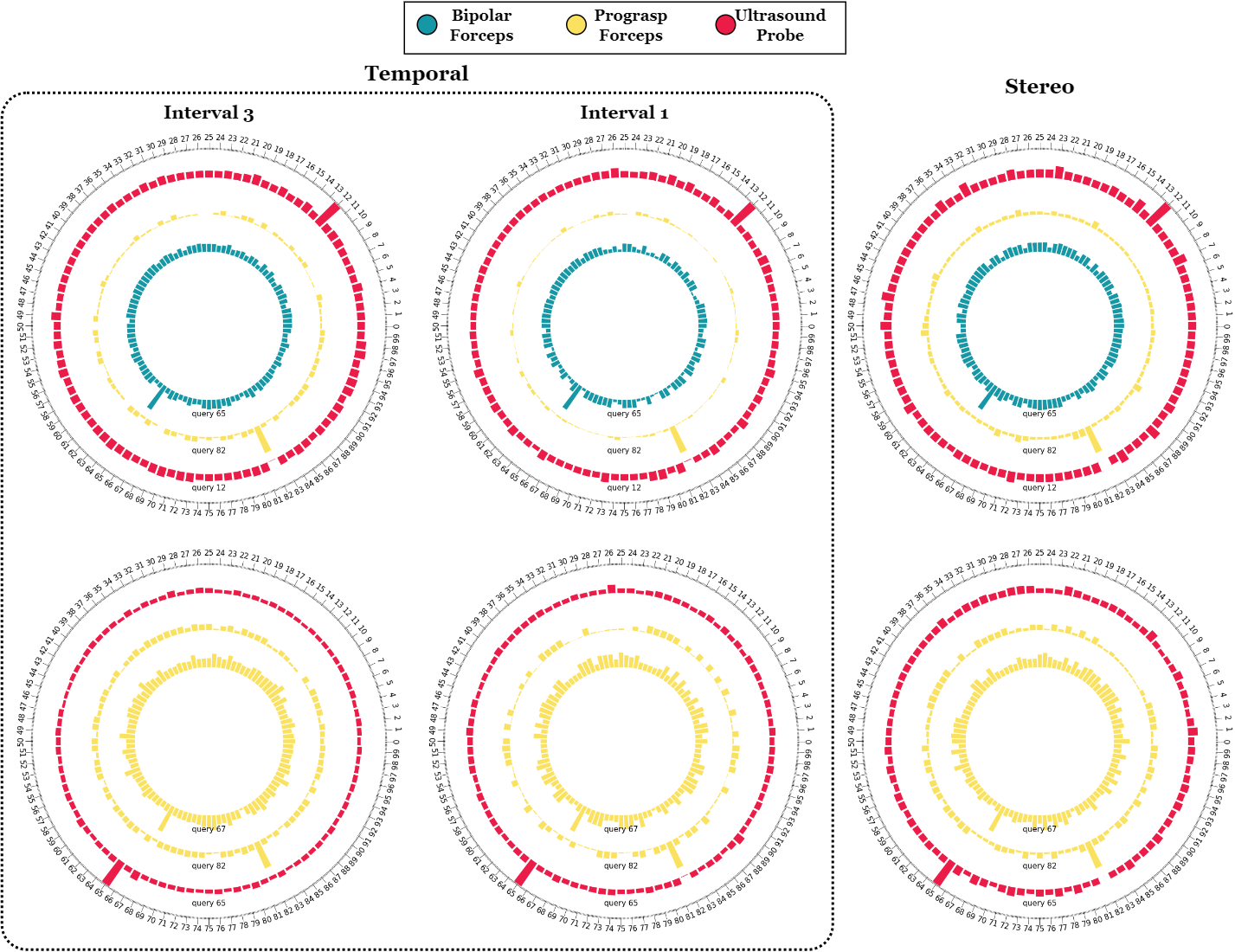}
	\caption{Quality Analysis of identity consistency in one stereo clip from embedding similarity perspectives. Every ring of circular chart represents the similarities between one valid object query of reference frame and 100 object queries of target frame. Valid object queries are ascertained by Hunguarian Matching. The ring color is set based on the category of valid object query.}
	\label{fig:draw_sequence_embedding}
\end{figure*}

\begin{figure*}[!ht]
	\centering
	\includegraphics[width=\textwidth]{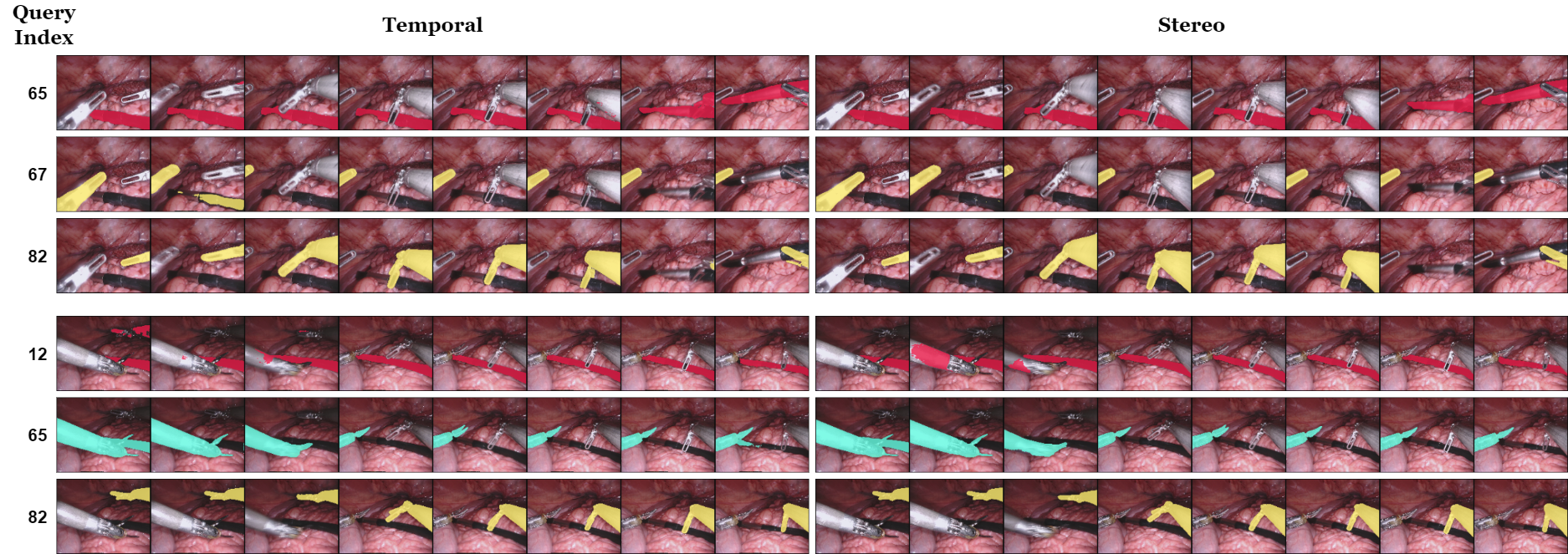}
	\caption{Quality Analysis of identity consistency in one stereo clip from query segmentation perspectives.}
	\label{fig:idc}
\end{figure*}

\textbf{Query Alignment Analysis.} As mentioned in~\ref{sect:qa}, it is assumed that object queries within a stereo clip, indexed by the same object query index, are assigned with the same identity ID after query alignment operation and the application of identity alignment loss constraints. This assumption serves to streamline the generation of tracklets without the need for video annotations. To validate the effectiveness of these mechanisms in preserving identity consistency, we present the graphical relationship between query indexes and identity IDs through a stereo clip as illustrated in Figure~\ref{fig:sequecnce_identity}. Distinct rows are indicative of different examples, whereas columns represent different methodologies. For each circular chart, the eight rings from outermost to innermost correspond to eight continuous frames, with 100 query indexes distributed along the counter-clockwise direction of each ring. The items highlighted in color indicate object queries that correspond to specific instances, whereas the gray items represent non-object queries. The corresponding color is set based on the category of matched instance. If \textbf{the same instance is aligned with same query index across different frames}), these mechanisms can fulfill the assumption. For comparison, we visualize baseline, our proposed method with only identity alignment loss constraints and full LACOSTE. The first two columns of Figure~\ref{fig:sequecnce_identity} show that only introducing the identity alignment loss can enhance the identity consistency relative to baseline. The query alignment mechanism deepens the above advantage as illustrated in the last two columns. However, it it worth noting that two mechanisms of LACOSTE are more suitable for short clips rather long videos. For fast motion and severe deformation circumstances, the methods are more inclined to find another new query index aligned with identity like tracking methods. The examples of Figure~\ref{fig:sequecnce_identity} keep the same with those of Figure~\ref{fig:draw_sequence_embedding} and Figure~\ref{fig:idc}.

\textbf{Embedding Similarity Analysis.} To illustrate the identity consistency encoded in object queries more effectively, we visualize query embedding similarities across both temporal and stereo frames as shown in Figure~\ref{fig:draw_sequence_embedding}. Rows denote examples while columns respectively represent interval-3,inter-1 temporal frames and stereo frame. For every ring of circular chart, a valid object query of current frame is analyzed for similarities with all 100 object queries from temporal or stereo frame. If the similarity value between object queries with the same index is higher, the method achieves greater identity consistency. In particular, for the first row example, the valid object queries of the current frame are the $12_{th}$,$65_{th}$, and $82_{nd}$ as indicated in the LACOSTE column of Figure~\ref{fig:sequecnce_identity}. As for query $65_{th}$ ring , the peak similarities are concentrated at the $65_{th}$ bar in all charts of first row, thereby preserving identity consistency. Analogous situations can be observed in other instances. We also visualize the segmentation results for each valid object query through a stereo clip to present the identity consistency explicitly as shown in Figure~\ref{fig:idc}.

\begin{table}[!htbp]
\begin{center}
    \caption{Evaluation on temporal consistency of segmentation results.}
		\begin{tabular}{lccccc}
            \hline
			\textbf{Method} & & \multicolumn{4}{c}{\textbf{\evdb}}\\
            \cline{3-6}
            \textbf{} & & \textbf{Ch}\_\iou & $\mathcal{J}$ & $\mathcal{F}$ & $\mathcal{J\&F}$\\
			\hline
            XMEM        & & 70.2 & 72.3 & 72.0 & 72.2 \\
            XMEM(F)     & & 72.6 & 73.1 & 73.6 & 73.4 \\
			SAM2        & & 82.6 & 81.4 & 81.5 & 81.4 \\
            SurgSAM-2   & & 84.4 & 84.3 & 84.0 & 84.1 \\
            SurgSAM-2(P) & & 84.1 & 84.2 & 83.8 & 84.0 \\
			LACOSTE(S)  & & \textbf{85.2} & \textbf{85.8} & \textbf{85.3} & \textbf{85.5} \\
            \hline
		\end{tabular}
	\label{tab:SAM2VOS}
\end{center}
\vspace{-5mm}
\end{table}

\textbf{Temporal Consistency of Segmentation Results.}
 We further analyze our temporal consistency of segmentation results together with some video object segmentation (VOS) methods including SAM2~\citep{ravi2024sam,shen2024performance} and SurgSAM-2~\citep{liu2024surgical} which is based on SAM2 and fine-tuned with \evdb. We also supplement fine-tuned results of XMEM~\citep{cheng2022xmem} with the same dataset, represented as XMEM(F). Actually, ours and VOS paradigms are inherently tailored for distinct tasks with different challenges. It is important to clarify that our intention is not to assert a superior temporal consistency compared to VOS methods. Instead, we aim to offer researchers potential insights and \textbf{feasible} alternatives for establishing temporal correlations in scenarios where video-level annotations are unavailable. The temporal consistency of segmentation results, which is not the primary focus or contribution of this paper, is an inherited outcome by mitigating misclassification and improving segmentation effectiveness. Particularly, VOS methods receive the annotated first frame for each object of the video sequence and propagate them through time dimension. For the LACOSTE model, we maintain the original inference process without utilizing any annotation. We present several VOS metrics, including Jaccard index $\mathcal{J}$, contour accuracy $\mathcal{F}$, and their average $\mathcal{J}\&\mathcal{F}$, together with semantic segmentation metric \textbf{Ch}\_\iou in Table~\ref{tab:SAM2VOS}. We use category IDs as object IDs in these experiments. For all metrics, LACOSTE achieves comparable or favorable results with the other methods. In fact, our method does not require any annotations during inference process while the competing methods require so. Additionally, we substitute the annotations of reference frames with our predictions for SurgSAM-2 to validate the cooperative potential, denoted as SurgSAM-2(P). Our approach can deliver semantic prompts for category-agnostic SAM2-based VOS methods when reference annotations are unavailable.

\section{Conclusion}
In this study, we systematically explore temporal information and stereo cue in surgical instrument segmentation tasks. LACOSTE extends common query-based segmentation methods with proposed disparity-guided feature propagation module, stereo-temporal set classifier and location-agnostic classifier to mitigate surgical domain challenges. Exhaustive experiments have been conducted on the benchmark robot-assisted surgery datasets.Our method generalizes well on all benchmarks and achieves comparable or favorable results with previous state-of-the-art approaches. We conclude that introducing temporal and stereo information improves the results for applications involving complicated classification in different circumstances. The proposed framework can be helpful to downstream applications that depend on tool identification and segmentation. We hope that our analysis and the innovations to mitigate the challenges specific to surgical instruments will spark similar interests in introducing specific information and attributes in other domains.
\bibliographystyle{model2-names.bst}\biboptions{authoryear}
\bibliography{refer}
\end{document}